\documentclass[runningheads]{llncs}
\usepackage{graphicx}

\usepackage{tikz}
\usepackage{comment}
\usepackage{amsmath,amssymb} 
\usepackage{color}
\usepackage{ulem}
\usepackage{cite}
\usepackage{multirow}
\usepackage{algpseudocode}
\usepackage{multicol}
\usepackage{algorithm}
\usepackage[tracking=true]{microtype}
\usepackage{verbatim}
\usepackage{authblk}
\usepackage[symbol*]{footmisc}
\usepackage{wrapfig}

\algrenewcommand\alglinenumber[1]{\scriptsize #1:}
\floatname{algorithm}{Algorithm}

\usepackage[accsupp]{axessibility}  
\usepackage[pagebackref,breaklinks,colorlinks]{hyperref}

\DefineFNsymbolsTM{otherfnsymbols}{%
  \textasteriskcentered *
  \textdagger    \dagger
  \textbullet \circ
  \textsection   \mathsection
  \textdaggerdbl \ddagger
  \textbardbl    \|%
  \textparagraph \mathparagraph
}%

\setfnsymbol{otherfnsymbols}

\begin{document}
\pagestyle{headings}
\mainmatter
\def\ECCVSubNumber{744}  

\title{Streamable Neural Fields} 

\titlerunning{Streamable Neural Fields}
%

\newcommand\CoAuthorMark{\footnotemark[\arabic{footnote}]}
\author{Junwoo Cho\inst{1}\thanks{Equal contribution, alphabetically ordered.} \and
Seungtae Nam\inst{1}\protect\CoAuthorMark \and
Daniel Rho\inst{1} \and
Jong Hwan Ko\inst{1, 2} \and
Eunbyung Park\inst{1, 2}\thanks{Corresponding author.}}

\authorrunning{J. Cho, S. Nam, D. Rho, JH. Ko and E. Park}
%
\institute{Department of Artificial Intelligence, Sungkyunkwan University\\ \and
Department of Electrical and Computer Engineering, Sungkyunkwan University\\
\email{\{jwcho000,stnamjef,daniel231,jhko,epark\}@skku.edu}}
\maketitle

\begin{abstract}
Neural fields have emerged as a new data representation paradigm and have shown remarkable success in various signal representations. Since they preserve signals in their network parameters, the data transfer by sending and receiving the entire model parameters prevents this emerging technology from being used in many practical scenarios.
We propose streamable neural fields, a single model that consists of executable sub-networks of various widths. 
The proposed architectural and training techniques enable a single network to be streamable over time and reconstruct different qualities and parts of signals.
For example, a smaller sub-network produces smooth and low-frequency signals, while a larger sub-network can represent fine details.
Experimental results have shown the effectiveness of our method in various domains, such as 2D images, videos, and 3D signed distance functions.
Finally, we demonstrate that our proposed method improves training stability, by exploiting parameter sharing.
Our code is available at \url{https://github.com/jwcho5576/streamable_nf}.

\end{abstract}

\section{Introduction}
Neural fields~\cite{xie2021neural} have emerged as a powerful representation of real-world signals. It uses a multilayer perceptron (MLP) that takes inputs as the spatial or temporal coordinates and produces signal values in arbitrary resolutions. Thanks to recent advances such as input feature encoding~\cite{zhong2020reconstructing, mildenhall2020nerf, tancik2020fourier} and periodic activation functions~\cite{sitzmann2020implicit}, it can faithfully reconstruct complex and high-frequency signals. It has achieved great success in various signal representations such as images~\cite{chen2021learning, martel2021acorn, muller2022instant}, 3D shapes~\cite{park2019deepsdf, mescheder2019occupancy, chen2019learning, chabra2020deep, erler2020points2surf, genova2019learning, jiang2020local, sitzmann2020metasdf}, and novel view synthesis~\cite{mildenhall2020nerf, zhang2020nerf++, martin2021nerf, schwarz2020graf, chan2021pi, niemeyer2021giraffe, lin2021barf}.

There are still many challenges that prevent this emerging technique from being used in practical scenarios.
In neural fields, the network itself is a data representation~\cite{dupont2022data} (the signals are stored as the parameters of neural networks), and the signal transmissions are done by sending and receiving the entire model parameters.
Thus, finding the optimal model size is crucial for lower latency and higher throughput. 
A naive approach would train different size networks multiple times by increasing depths and widths. However, it would not be an affordable solution since training even a single network takes a long time to converge for deep neural networks.
It is tempting to predetermine various network configurations (e.g., widths and depths) for different sizes and types of signals, yet it is not a feasible solution either since the required size of the parameters is determined by the complexity of the signals, not the size or type of the data.

\begin{figure*}[t]
  \includegraphics[width=\textwidth]{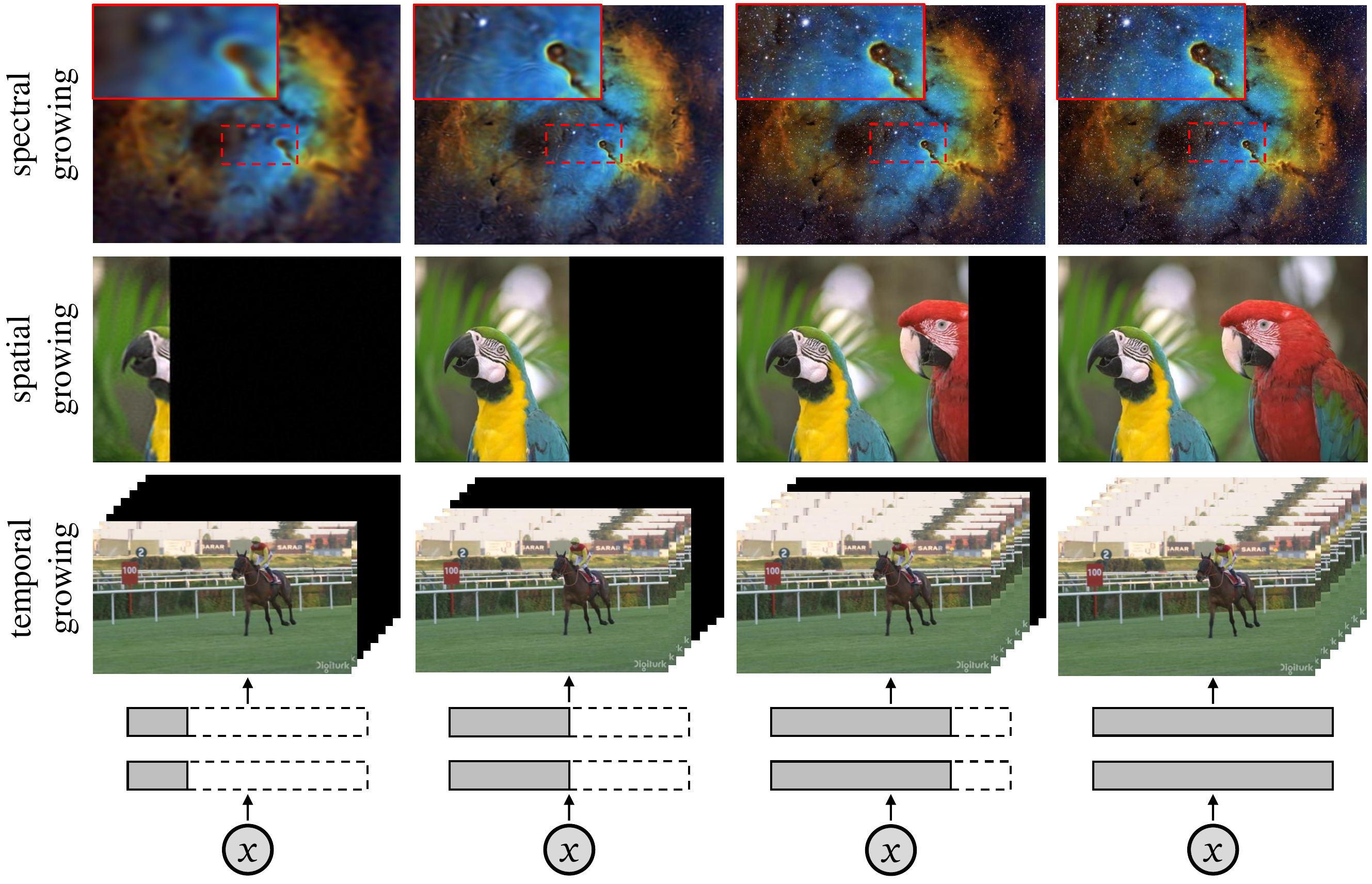}
  \caption{Representing a varying signal with streamable neural fields.
  It is a \textit{single} neural network executable at different widths that can reconstruct varying signal domains.
  \textit{Spectral growing}: larger network reconstructs more high frequency details. 
  \textit{Spatial growing}: larger network reconstructs more pixel locations. 
  \textit{Temporal growing}: larger network reconstructs more video frames.
  }
  \label{fig:main_fig}
\end{figure*}

In addition, the raw signals often need to be transmitted in different resolutions or qualities.
For example, in media streaming services, users want to receive various quality signals according to their circumstances, e.g., high-resolution videos at home and lower quality on mobile devices.
Real-time encoding on-demand is not viable since it requires a long latency gradient descent on deep neural networks.
As an alternative approach, we could locally store multiple sizes of the networks representing different qualities of the signals beforehand.
However, it is a waste of storage space and not an acceptable solution given the exponential growth of media data.

Unlike most of the standard compression algorithms, such as JPEG~\cite{pennebaker1992jpeg} and MPEG~\cite{le1991mpeg} which are designed to be easily broken down into smaller pieces for potential use cases, including streaming service or partial reconstruction in poor network connections, a neural field cannot be decoupled into meaningful chunks.
All the weight parameters are highly intertwined, and missing a small part of them would result in catastrophic failures for signal reconstruction.

We propose \textit{streamable neural fields} to overcome the issues mentioned above.
We suggest training techniques and architectural designs that enable a \textit{single} trained network to be separated into executable sub-networks of various widths.
With \textit{single training} procedure, the proposed algorithm can generate parameters of a single network that are \textit{streamable over time} and capable of reconstructing \textit{various qualities of signals} (Fig.~\ref{fig:main_fig}).
Each sub-network is in charge of representing some portion of the signals. For example, a small sub-network can only generate the signals in specific quality or a specific temporal (or spatial) range of the signals.
A wider network that subsumes narrower sub-networks can represent the additional signals that are not encoded in the narrower sub-networks.
By streaming the network parameters (from narrower to wider sub-networks), the signals will be progressively reconstructed in visual quality and temporal (or spatial) orders, which is desirable in many useful scenarios.

In sum, we present a single neural network that can represent multiple visual qualities and spatial (or temporal) ranges and decode signals in a streamline. The proposed network architecture and training strategy maximize the use of learned partial signals preserved in the small sub-networks. The larger networks explicitly make use of them, resulting in a more stable training procedure, improving reconstruction performance, and increasing parameter efficiency. We show the proposed method's effectiveness in various signals, including images, videos, and 3D shapes.

\section{Related Work}
\subsubsection{Neural fields and spectral bias}
Neural fields, also known as coordinate-based neural representations or implicit neural representations have shown great success in representing natural signals such as images~\cite{chen2021learning, martel2021acorn, muller2022instant}, videos~\cite{sitzmann2020implicit, li2021neural}, audios~\cite{sitzmann2020implicit}, 3D shapes~\cite{park2019deepsdf, mescheder2019occupancy, chen2019learning, chabra2020deep, erler2020points2surf, genova2019learning, jiang2020local, sitzmann2020metasdf} and view synthesis~\cite{mildenhall2020nerf, zhang2020nerf++, martin2021nerf, schwarz2020graf, chan2021pi, niemeyer2021giraffe, lin2021barf}.
They struggled to represent high-frequency details due to low-dimensional inputs and spectral bias in training procedure~\cite{tancik2020fourier, rahaman2019spectral}. 
Fourier feature encodings~\cite{tancik2020fourier,mildenhall2020nerf} and periodic non-linear activation function~\cite{sitzmann2020implicit}
enabled networks to represent fine details and have been successful.
Although spectral bias is an unpleasant training behavior in many practical tasks, our work exploits this phenomenon to implement a neural field that can decode signals in various qualities with a single neural network.

\subsubsection{Learning decomposed signals}
Several studies on neural fields represent spatially partitioned signals using voxel grids ~\cite{yu2021plenoxels, yu2021plenoctrees, sun2022direct}, latent codes ~\cite{chen2021learning, mehta2021modulated}, and a group of neural networks ~\cite{reiser2021kilonerf}.
Voxel-based approaches~\cite{yu2021plenoxels, yu2021plenoctrees, sun2022direct} directly bake a radiance field into the feature grids.
The feature grids can be transmitted in a streamline, but their size is very large, which is unfavorable in streaming and compression.
Another line of works~\cite{chen2021learning, mehta2021modulated} divides an image into tiles and encodes them as latent vectors.
While the latent vectors are much smaller than the feature grids, a decoder is required on the client-side.
Our method allows the network parameters to be streamed instead of the latent vectors and does not require an additional decoder to reconstruct the signal.
KiloNeRF~\cite{reiser2021kilonerf} subdivides a single network into numerous tiny ones.
Similar to our method, the parameters are streamable.
However, thousands of networks should be trained independently.
In addition, we also found that learning a spatially partitioned scene with independent models yields line artifacts (Fig.~\ref{fig:glitch}), while our method seamlessly reconstructed the scenes.


On the other branch of work, a signal is partitioned in a frequency domain and learned hierarchically.
Takikawa et al. \cite{takikawa2022variable} proposed to learn multiresolution codebooks similar to~\cite{takikawa2021neural}, allowing variable bitrate streaming.
In a similar spirit to ours, recent works \cite{lindell2022bacon, shekarforoush2022residual, landgraf2022pins} have suggested a single network representing a signal with various bandwidths.
Input layers are laterally connected to each intermediate layer, and the intermediate~\cite{landgraf2022pins} (or additional output~\cite{lindell2022bacon, shekarforoush2022residual}) layers reconstruct band-limited signals.
To constrain the bandwidth, \cite{lindell2022bacon, shekarforoush2022residual} initialize and fix the parameters of each input layer such that they are uniformly distributed over a certain frequency range.
Progressive implicit networks (PINs)~\cite{landgraf2022pins} sort sampled frequencies in ascending order, divide them into subsets, and use each subset as the Fourier encoding~\cite{mildenhall2020nerf, tancik2020fourier} frequencies.
Although these works~\cite{lindell2022bacon, shekarforoush2022residual, landgraf2022pins} share some similarities with our method, there are significant differences. 
First, we exploit the spectral bias to learn the optimal frequency bandwidths given the limited network capacity while \cite{lindell2022bacon, shekarforoush2022residual} manually constrains the bandwidths, which may result in inefficient use of the network capacity. 
Second, our method is agnostic to input encoding methods while \cite{landgraf2022pins} designed customized algorithms for particular input encoding methods. Finally, we also proposed to grow widths of networks instead of depths~\cite{lindell2022bacon, shekarforoush2022residual, landgraf2022pins}.

\subsubsection{Dynamic neural networks}
Unlike traditional neural networks with static architecture and size, dynamic neural networks can expand or shrink in size on the fly during training and inference.
They can adapt to various computing environments and achieve a trade-off between efficiency and accuracy.
One branch of dynamic neural networks related to our work is slimmable neural networks (SNN)~\cite{yu2019universally, yu2019slimmable}, which dynamically expands the channel width of convolution filters during training.
In SNN, without re-training each different network architecture, the model is executable for multiple predefined widths.
Each trained sub-network has similar or better performance than individually trained models with the help of knowledge distillation~\cite{hinton2015distilling} and parameter sharing.

In the field of lifelong learning (or incremental learning)~\cite{thrun1994lifelong}, neural networks learn from a sequence of multiple tasks.
Progressive neural network~\cite{rusu2016progressive} dynamically expands while transferring knowledge from prior tasks to new ones to handle more tasks and overcome catastrophic forgetting behavior~\cite{mccloskey1989catastrophic, goodfellow2013empirical}.
In this work, we focus on representing signals and propose a neural field that can dynamically grow network size to represent a higher-quality or broader range of signals while preserving the representations from small sub-networks.


\begin{figure*}[t]
  \includegraphics[width=\textwidth]{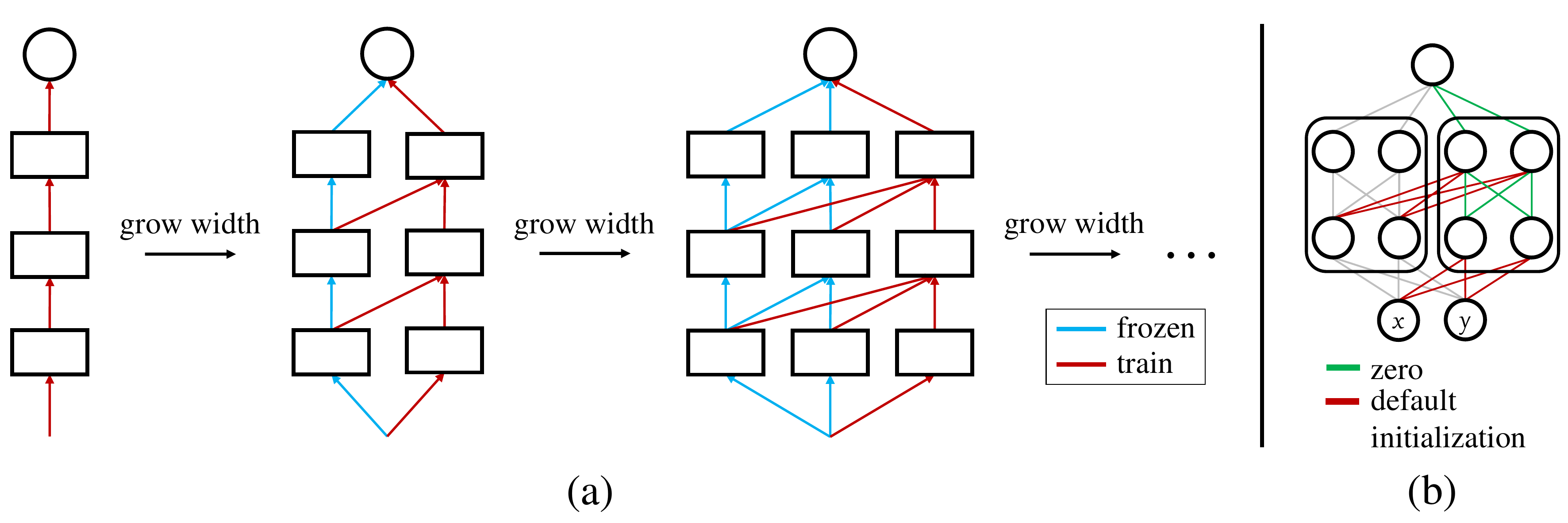}
  \vspace*{-8mm}
  \caption{(a) Network architecture and training procedure of streamable neural fields.
  When training for each sub-network is done, it can grow the width into arbitrary size.
  To keep the output of each trained sub-network, previous weights are kept frozen.
  (b) Parameter initialization for newly added weights.
  Leaving the lateral connections (red) to default initialization, the remaining weights (green) are initialized to zero for fast convergence.
  We used SIREN initialization~\cite{sitzmann2020implicit} as default.}
  \vspace*{-2mm}
  \label{fig:training_procedure}
\end{figure*}

\section{Streamable Neural Fields}
\vspace*{-1.5mm}

This section explains training techniques and architectural designs of \textit{streamable neural fields} that consist of executable sub-networks of various widths. Once training is completed, a single network can present signals in various qualities without re-training (spectral growing). Narrower sub-networks preserve low-frequency signals and wider sub-networks contain high-frequency details. This functionality can support many practical applications by selecting a width to offer requested quality signals on demand. In general, there is no straightforward solution to manually divide weights to achieve this goal since the network parameters are highly entangled.

In addition, streamable neural fields also support spatial and temporal growing. Each runnable sub-network can only represent a specific part of the entire signal. As depicted in Fig.~\ref{fig:main_fig}, one would expect to send or receive a video sequentially over time, and one would prefer to receive a small part of a large image.
It would be beneficial for transmitting large-size signals, such as high-resolution images or videos.
\vspace*{-1mm}
\subsection{Network architecture and progressive training}
\vspace*{-1.5mm}

Our network architecture and training procedure are illustrated in Fig.~\ref{fig:training_procedure} (a). The model starts training with a small and narrow MLP to predict a target signal. Once converged, it grows its width by arbitrary size. Similar to progressive neural networks architecture~\cite{rusu2016progressive}, we remove the weights that connect from newly added hidden units to the previous units, which prevents the added units from affecting the small network's output. We also freeze weights in the small network and only update the newly added network parameters. This progressive training strategy encourages the large network to use the knowledge learned by the previous small network and only learn the residual signals that the small network cannot capture. We keep iterating this process until the desired signal quality or spatial/temporal size is fulfilled.

\subsubsection{Progressive training vs slimmable training}

Our general purpose is to create a single network that is executable at various widths.
We found that the training technique in slimmable networks ~\cite{yu2019slimmable} can also achieve the goal for image and video fitting tasks.
Unlike the proposed progressive training, it iterates over predefined widths, takes a sub-network of the corresponding width, and computes loss using the target signal prediction. 
The gradients of the sub-networks are accumulated until it visits every width, and weights are updated all at once.
Experimental results have shown that progressive training outperforms slimmable training in reconstruction quality and convergence speed. One possible explanation is that the target residual signals for wider networks change over the training course, which results in slowing down convergence speed. 
The details of the progressive and slimmable training algorithms are described in Alg.~\ref{alg:progressive} and Alg.~\ref{alg:slimmable} respectively.

\begin{figure}[t]
\begin{multicols}{2}
  \scriptsize \begin{algorithm}[H]
    \caption{Progressive training}\label{alg:progressive}
    \begin{algorithmic}[1]
    \Require Inputs $x$, targets $y$
    \State $\theta=\{ \}$
    \While {not done}
    \State $\theta_\text{new} \leftarrow $ GrowNetwork()
    \State Initialize $\theta_\text{new}$
    \State $\theta \leftarrow \theta \cup \{\theta_\text{new}\}$
    \For{epoch = 0 to $n_{steps}$}
        \State Predict signal values $\hat{y}=f_\theta(x)$
        \State Compute loss $\mathcal{L}(y, \hat{y})$
        \State Compute gradients $\nabla_{\theta_\text{new}}\mathcal{L}$
        \State Update $\theta_\text{new}$
    \EndFor
    \EndWhile
    \end{algorithmic}
  \end{algorithm}

  \columnbreak

  \begin{algorithm}[H]
    \caption{Slimmable training}\label{alg:slimmable}
    \begin{algorithmic}[1]
    \Require Inputs $x$, targets $y$
    \Require Parameters $\{\theta_{w_1}, \dots, \theta_{w_K}\}$
    \For{epoch = 0 to $n_{steps}$}
    \State $\theta=\{ \}$
    \For{$i=1$ to $K$}
        \State $\theta \leftarrow \theta \cup \{\theta_{w_i}\}$
        \State Predict signal values $\hat{y}=f_\theta(x)$
        \State Compute loss $\mathcal{L}(y, \hat{y})$
        \State Compute gradients $\nabla_\theta \mathcal{L}$
        \State Accumulate gradients
    \EndFor
    \State Update $\theta_{w_i}, \forall i \in \{1,...,K\}$
    \EndFor
    \end{algorithmic}
  \end{algorithm}
\end{multicols}
\end{figure}

\subsubsection{Training loss}
In every task, the training objective is to minimize the mean squared error (MSE) between prediction and ground truth target value:
\begin{align}
    \min_{\theta}\frac{1}{N}\sum_{i=1}^{N}\Vert f_\theta (x_i)-y_i\Vert_{2}^{2},
\end{align}
where $\{(x_i, y_i)\}_{i=1}^{N}$ is coordinate and corresponding signal value pair, e.g., $x_i \in \mathbb{R}^2$ and $y_i \in \mathbb{R}^3$ for image signals, e.g., RGB colors. $f_\theta$ is a neural network parameterized by $\theta$.
For spectral growing, the target signal is fixed to $y_i$.
On the other hand, for spatial and temporal growing, we divided the input coordinate and the ground truth signal into the desired size for each network to represent.
For example, to only represent a specific part of a certain sub-network, our objective becomes:
\begin{align}
    \min_{\theta}\frac{1}{N}\left(\sum_{i\notin S}\Vert f_\theta (x_i)\Vert_{2}^{2} + \sum_{i \in S}\Vert f_\theta (x_i)-y_i\Vert_{2}^{2}\right),
    \label{eq:spatial_growing}
\end{align}
where $S$ is the set of indices that belongs to the part. We train the network to predict zero for coordinate locations that do not belong to the part of interest. We empirically found that the first term in Eq.~\ref{eq:spatial_growing} is critical to seamlessly reconstructing the entire signals. The network usually predicts garbage outputs for the locations that are not part of training coordinates. It causes severe artifacts in the boundary areas when we stitch different parts of signals.



\subsubsection{Initialization}
A careful initialization is required when using periodic activation, in order to achieve high performance and convergence speed. Fig.~\ref{fig:training_procedure} (b) shows the suggested initialization scheme. We used the initialization method in SIREN~\cite{sitzmann2020implicit} for lateral connections: $w\sim \mathcal{U}(-\sqrt{6/n}, \sqrt{6/n})$, where $n$ is the number of input neurons in a particular layer. It helps the network maintain the activation distribution throughout layers. The remaining weights that connect between newly added neurons are set to zero. In a consequence, only information from the small sub-networks flows at the beginning of the training process. It would encourage the large network to utilize the small networks' knowledge first and keep newly added parameters from learning redundant signals already learned by the small networks. We empirically found that this technique increases network parameter efficiency and improves the convergence speed.

\subsection{Spectral decomposition of streamable neural fields}
In this section, we analyze a trained streamable neural field through the lens of spectral bias~\cite{rahaman2019spectral,tancik2020fourier}.
In neural fields using a simple MLP, the final output is a weighted sum of the previous layer's outputs (assuming no activation functions on the output layer).
Any signals can be represented as a sum of different frequency components, and it is theoretically possible to assign values to the weight matrix in the final layer in a way that the final output can progressively represent higher frequency components as increasing the width.

\begin{figure*}[t]
  \includegraphics[width=\textwidth]{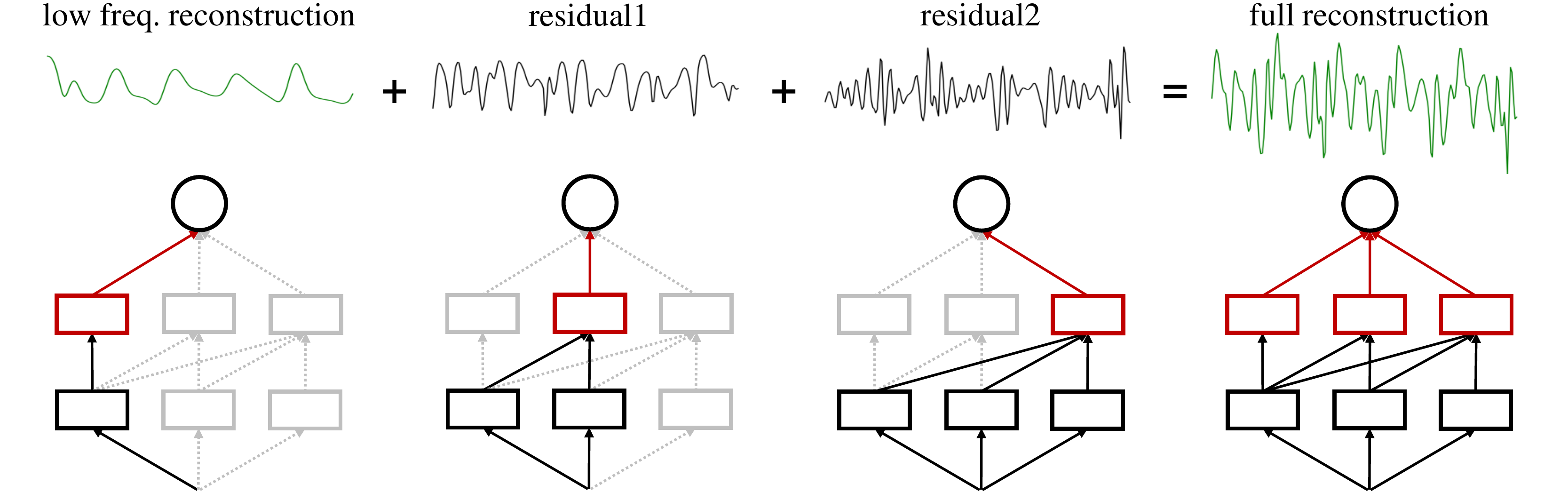}
  \caption{Spectral growing and residual representation of streamable neural fields.
  Due to spectral bias, a neural network trained with a standard MSE loss would naturally build up signals in increasing frequency orders.
  Narrower sub-networks will first learn low-frequency components and wider sub-networks will increasingly represent higher frequency signals as long as the capacity of the network permit.
  Pruning the partial weights of the final hidden layer outputs residual signals and linearly combining them (red boxes) gives the full reconstruction.}
  \label{fig:residual_representation}
\end{figure*}

The proposed progressive training scheme does not modify previously learned sub-network outputs, implying that the newly added hidden units only represent the residual signals (Fig.~\ref{fig:residual_representation}) and spectral bias~\cite{rahaman2019spectral,tancik2020fourier} claims that the network prioritizes learning low-frequency parts of signals.
Therefore, a narrow sub-network trained by the progressive training would represent low-frequency signal, and wide network will preserve high-frequency details.
The output value of a fully-connected neural network is a linear combination of the final hidden activation.
Our model progressively accumulates the final hidden activation learned by each sub-network.
More formally, we can formulate the output layer of an MLP as follows:
\begin{equation}
    \label{eq:aggregation}
    y = f_\theta(x) = \sum_{j=1}^{d}{w_j\phi_j(x)} = \underbrace{\sum_{j=1}^{s}{w_j\phi_j(x)}}_{\substack{\textrm{low frequency}\\ \textrm{reconstruction}}} + \underbrace{\sum_{k=s+1}^{d}{w_k\phi_k(x)}}_{\substack{\textrm{high frequency}\\ \textrm{residual}}},
\end{equation}
where $x \in \mathbb{R}^n$ is an $n$-dimensional coordinate, $y \in \mathbb{R}$ is a single-channel signal, $w \in \mathbb{R}^d$ is a weight vector in the final layer of the network, $\phi_j(x)$ is the $j$-th hidden unit in the final layer. The right-hand side in Eq.~\ref{eq:aggregation} is a decomposition into two partial sums, splitting the summation by index $s$ (width of a sub-network). We assume there is no bias term in the final layer.

We can interpret $\phi_j$ as basis functions and $w_j$ as coefficients. Unlike well-known basis functions, such as Fourier basis or Chebyshev polynomials, we learn basis functions through progressive training and do not have any constraints e.g., orthogonality and periodicity on a  Fourier basis. We solely rely on the inductive bias of MLP architecture and spectral bias of training neural networks to obtain spectral growing neural fields.

\subsubsection{Quality control}
The central premise of modern image compression algorithms is that the human eye cannot detect high-frequency components.
JPEG~\cite{pennebaker1992jpeg} uses discrete cosine transform (DCT) and quantization matrix to remove the high-frequency parts in an image that are not recognizable to human eyes.
It uses a quality factor that determines the compression rate and the reconstruction quality depending on the user's demand.
Our streamable neural fields gradually eliminate the high-frequency component as the width decreases.
Thus, choosing the width of the network to reconstruct a signal of desired quality is analogous to choosing a quality factor in image compression algorithms.

\begin{figure*}[t]
  \includegraphics[width=\textwidth]{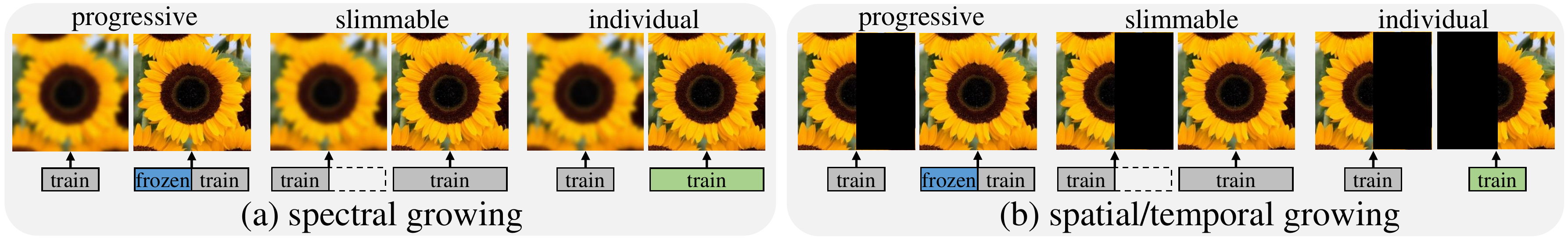}
  \caption{Description of our experimental setting.
  Every shown image above is the desired output signal and the number of parameters for each model is matched for a fair comparison.
  (a) Spectral growing experiment. The ground truth signal is fixed.
  (b) Spatial growing experiment. Individually trained models have constant network capacity.\\
  }
  \label{fig:baseline_description}
  \includegraphics[width=\textwidth]{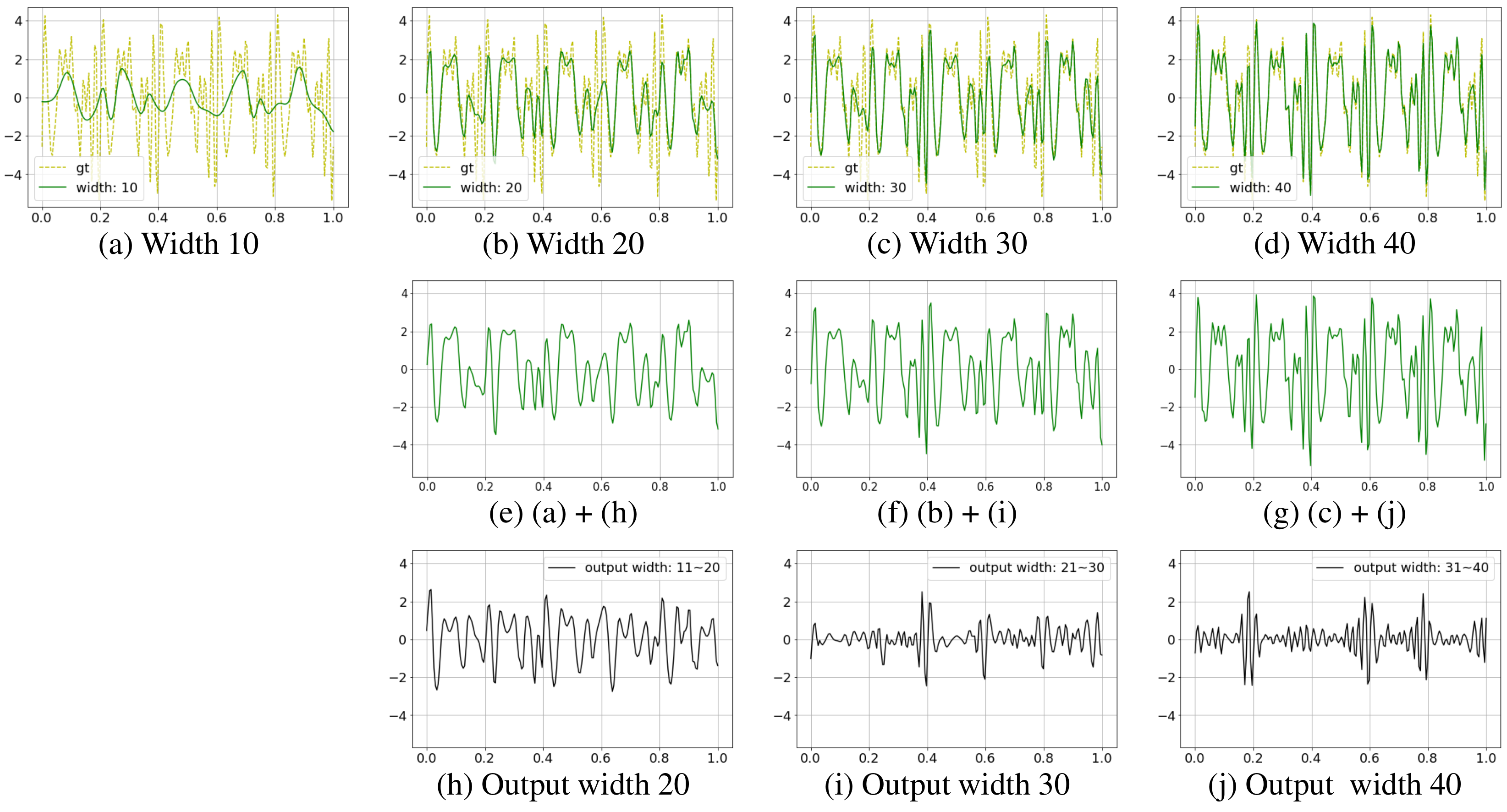}
  \caption{The results of the 1D sinusoidal function experiment. From (a) to (d): Output of each sub-network. As width grows, the model represents high-frequency details. Yellow dashed lines are the target ground truth signal. From (e) to (g): Summation of each sub-network and residual output. It is identical to the network outputs from (b) to (d). From (h) to (j): Residual output of each sub-network. The newly added weights reconstruct high-frequency residuals.}
  \label{fig:synthetic_result}
\end{figure*}

\section{Experiments}

We tested our model on various signal reconstruction tasks: 1D sinusoidal functions, 2D images, videos, and 3D signed distance functions (SDF).
Our experimental setting is described in Fig.~\ref{fig:baseline_description}.
For spectral growing, since there is no ground-truth residual signal, every individually trained model is trained to reconstruct the original target.
On the other hand, for spatial/temporal growing, we divided the ground truth signal into multiple patches/frames and make individual models that have identical network sizes.
Each model is then trained to represent specific image patches or video frames.
Note that streaming spatial/temporal growing individual models can also achieve sequential signal transmission, while spectral growing individual models don't.


\subsection{1D sinusoidal function reconstruction}
\label{sec:1d_recon}
In this section, we show the spectral growing of a signal by simple 1D scalar function fitting followed by~\cite{rahaman2019spectral}.
This experiment intuitively shows the residual representation of our model.
The target function is a mapping $f:[0, 1]\rightarrow \mathbb{R}$, constructed by summation of sinusoids with different frequencies and phase angles.
Starting with a width of 10 (the channel size of hidden layers), we gradually increased the size up to 40.
Each sub-network was trained for 150 epochs.
Fig.~\ref{fig:synthetic_result} shows the output signals and residuals learned by each sub-network.
As expected, the summation of low-frequency signals from small sub-networks and residual outputs give the same signal learned by larger sub-network.

\begin{figure*}[h]
  \includegraphics[width=\textwidth]{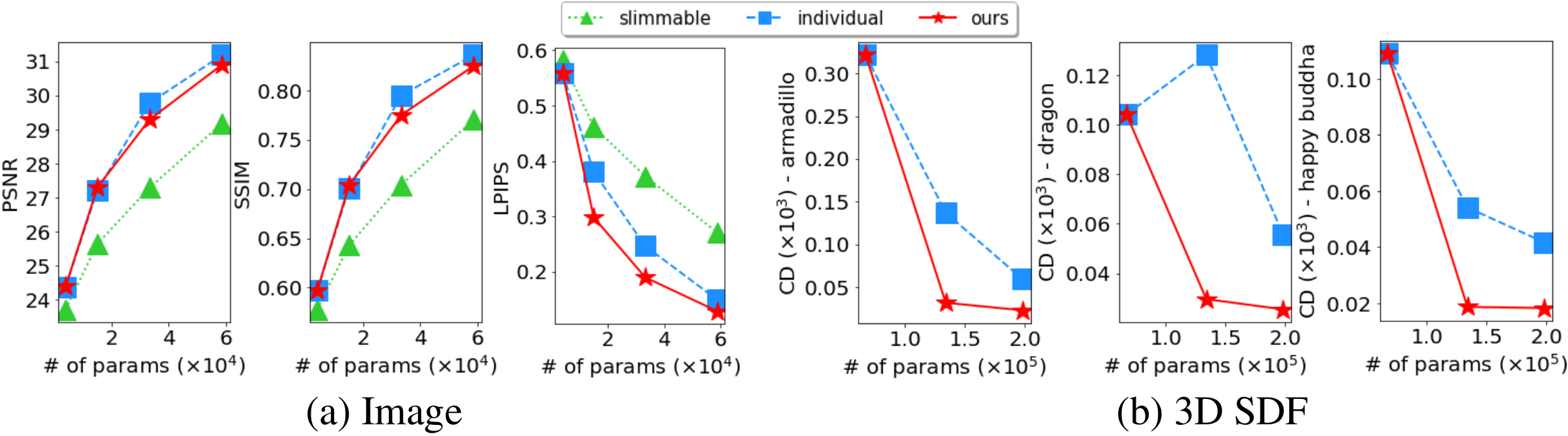}
  \caption{Quantitative result of the spectral growing experiment.
  (a): Averaged PSNR$\uparrow$, SSIM$\uparrow$, and LPIPS$\downarrow$ of 24 Kodak images.
  (b): Chamfer distance (multiplied by $10^3$) of three 3D shapes.}
  \label{fig:spectral_result}
\end{figure*}

\subsection{Spectral growing in images and 3D shapes}
\label{sec:spectral_exp}

\subsubsection{Images}
\label{sec:spectral_image}
We trained on 24 images in the Kodak dataset for spectral growing. 
The network size grows three times, using the same ground-truth images for four different executable sub-networks.
We compared to baseline method, denoted as \textit{individual}, which trains MLPs with different sizes. The number of hidden units in \textit{individual} models are adjusted to match the total number of parameters in sub-networks of the \textit{streamable} network. \textit{streamable (progressive)} denotes a streamable model trained with the proposed progressive training, and \textit{streamable (slimmable)} for the slimmable training.

Fig.~\ref{fig:spectral_result} (a) shows the averaged PSNR, SSIM~\cite{wang2004image} and LPIPS~\cite{zhang2018unreasonable}.
As expected, enlarging the network capacity gives a higher quality image.
We found that the progressive training strategy outperforms the slimmable training. 
We believe the fact that during the slimmable training, large sub-networks affect the outputs of the small sub-networks, and vice versa would hurt the final reconstruction performance and convergence speed.
In terms of the final reconstruction quality compared to \textit{individual}, our method performs comparably given the same number of parameters.
Especially, LPIPS gives a good score to \textit{streamable (progressive)} model, even though when PSNR and SSIM do not.
Since raw Kodak images contain undetectable high frequency components, our model gets rid of these due to spectral bias~\cite{rahaman2019spectral}.
Although our model does not \textit{exactly} reconstruct the target signal in terms of PSNR, low LPIPS implies that it sufficiently represents important features in terms of human's visual perception.

Moreover, the \textit{individual} model in spectral growing does not have a streamable functionality, which means that the \textit{individual} requires much more parameters to represent various qualities, e.g., four \textit{individual} models for four different qualities.
Table~\ref{table:param_efficiency} shows the memory comparison between the \textit{streamable (progressive)} model and the \textit{individual} model.
We trained on Kodak image 23 and divided it into 15 sub-networks that reconstruct 15 spectral growing images.
The memory requirement of our \textit{streamable (progressive)} model increases much more slowly compared to the \textit{individual}.
The difference between the two models becomes larger as more sub-networks are created.

\begin{table}
\centering
    \begin{tabular}{@{\hskip 0.03in}c@{\hskip 0.03in}|@{\hskip 0.03in}c@{\hskip 0.09in}c@{\hskip 0.09in}c@{\hskip 0.09in}c@{\hskip 0.09in}c@{\hskip 0.09in}c@{\hskip 0.09in}c@{\hskip 0.03in}}
    \hline
    PSNR & 25.2 & 25.2$\sim$27.8 & 25.2$\sim$29.8 & \ldots & 25.2$\sim$38.5 & 25.2$\sim$39.2 & 25.2$\sim$39.8\\
    \hline
    ours & 1.8K & 5.2K & 10.2K & \ldots & 148K & 171K & 195K \\ 
    individual & 1.8K & 6.5K\tiny{($\times$1.25)} & 15.0K\tiny{($\times$1.47)} & \ldots & 418K\tiny{($\times$2.82)} & 499K\tiny{($\times$2.92)} & 590K\tiny{($\times$3.03)} \\
    \hline
    \end{tabular}
    \caption{The total number of parameters required to cover the indicated range (PSNRs in the first row) of image quality.
    The second row shows the number of parameters of each \textit{sub-network} in our model to reconstruct the PSNR range above, and the rightmost value (195K) subsumes all the numbers left.
    The values in the third row are obtained by accumulating the number of parameters of \textit{individual} models.}
    \label{table:param_efficiency}
\end{table}

\begin{figure*}[t]
  \includegraphics[width=\textwidth]{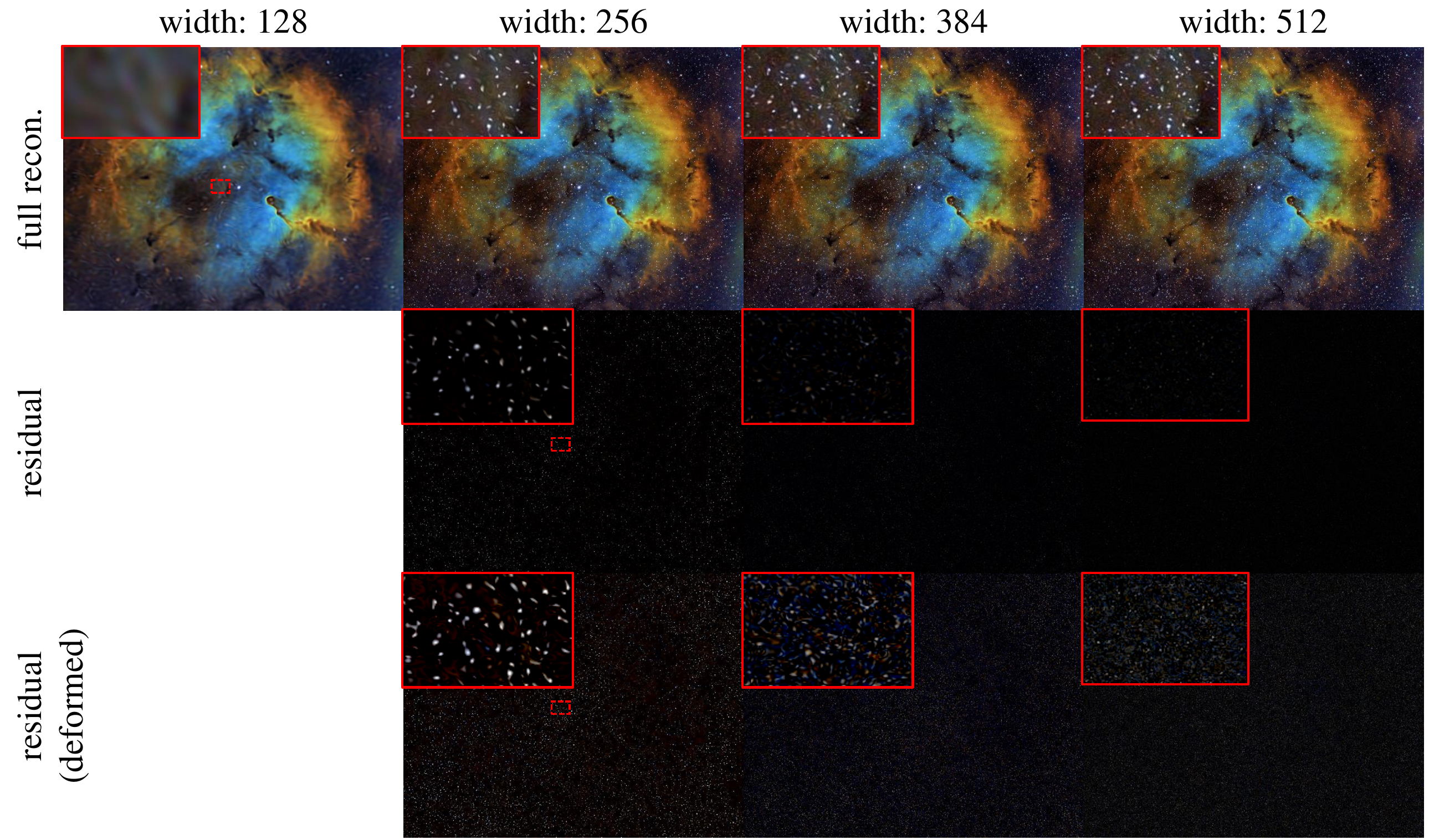}
  \caption{Visualizing residual representation of elephant's trunk nebula image (2560$\times$1984).
  As the network enlarges, darker stars gradually appear.
  The third row shows the deformed residual images for better visualization.
  Note that all images shown are the output of a single network.}
  \label{fig:nebula_qualitative_result}
\end{figure*}

In Fig.~\ref{fig:nebula_qualitative_result}, we show a qualitative result on a large size image. 
The original image presents many stars of various sizes and brightness.
After training each sub-network to the same target, we obtained residual signals by the method described in Fig.~\ref{fig:residual_representation}. 
The figure shows that the residual output images only contain small stars not captured by smaller sub-network.


\subsubsection{3D shapes}
\label{sec:spectral_sdf}
\begin{figure*}[t]
  \includegraphics[width=\textwidth]{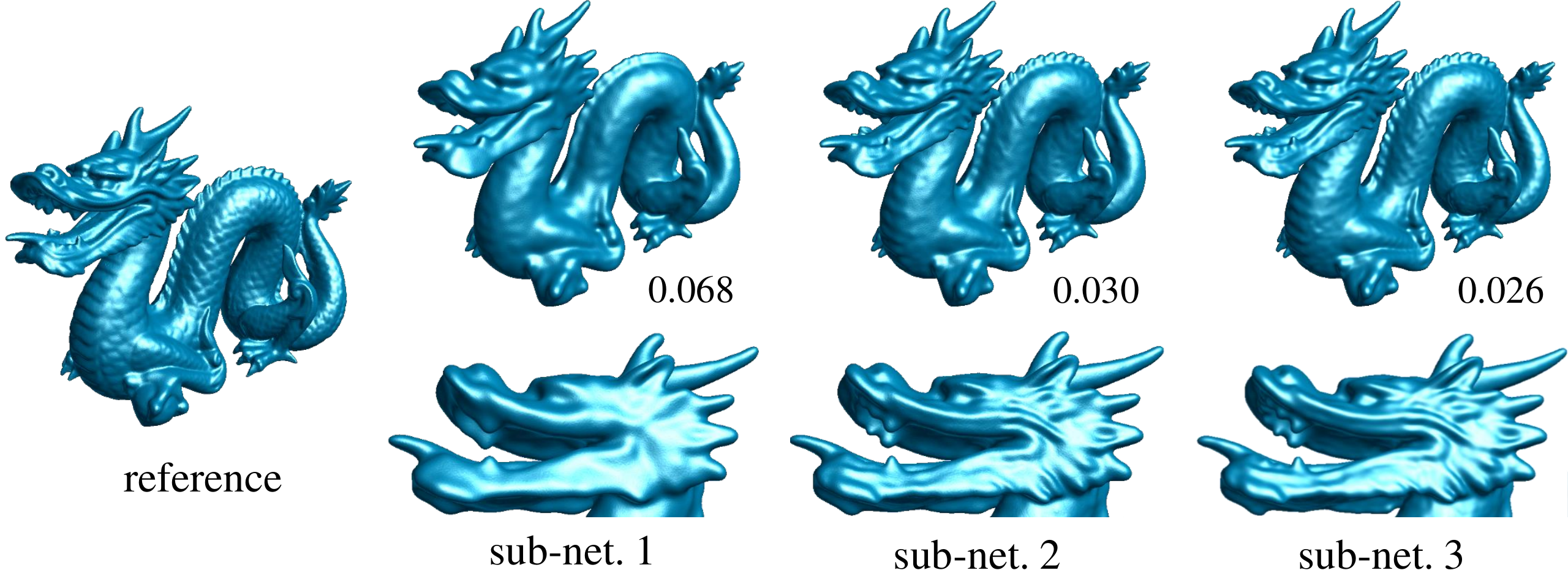}
  \caption{Spectral growing of \textit{Dragon} in Stanford 3D scanning repository.
  Indicated numbers are chamfer distances against the reference shape.}
  \label{fig:sdf_qualitative_result}
\end{figure*}

We also tested our method on 3D shapes represented by SDF.
We trained on \textit{Armadillo}, \textit{Dragon}, and \textit{Happy Buddha} from the Stanford 3D Scanning Repository. Following~\cite{park2019deepsdf}, we used bidirectional chamfer distance (CD) against the true shape to quantitatively compare different approaches.
The results are shown in Fig.~\ref{fig:spectral_result} (b) and the qualitative result of the \textit{Dragon} shape is illustrated in Fig.~\ref{fig:sdf_qualitative_result}.

Again, our single network outputs the same shape in various qualities. 
Despite many hyperparameter searches, the slimmable training approach did not converge on 3D shapes, so we only compared against the \textit{individual}. 
Even with the streamable functionality in our method, \textit{streamable (progressive)} shows better reconstruction quality given the same number of parameters. 
Again, \textit{individual} needs multiple independent models to support various quality outputs, which requires enormous storage space for 3D shapes.

In 3D shapes, training vanilla SIREN~\cite{sitzmann2020implicit} was very unstable, so some models fail on reconstruction (see Fig.~\ref{fig:spectral_result} (b) second column blue line).
However, our streamable model shows a steady performance improvement when the model capacity increases.
We provide an analysis of this phenomenon in terms of stable training behavior of the proposed progressive training (section~\ref{stable}).

\begin{figure*}[t]
  \includegraphics[width=\textwidth]{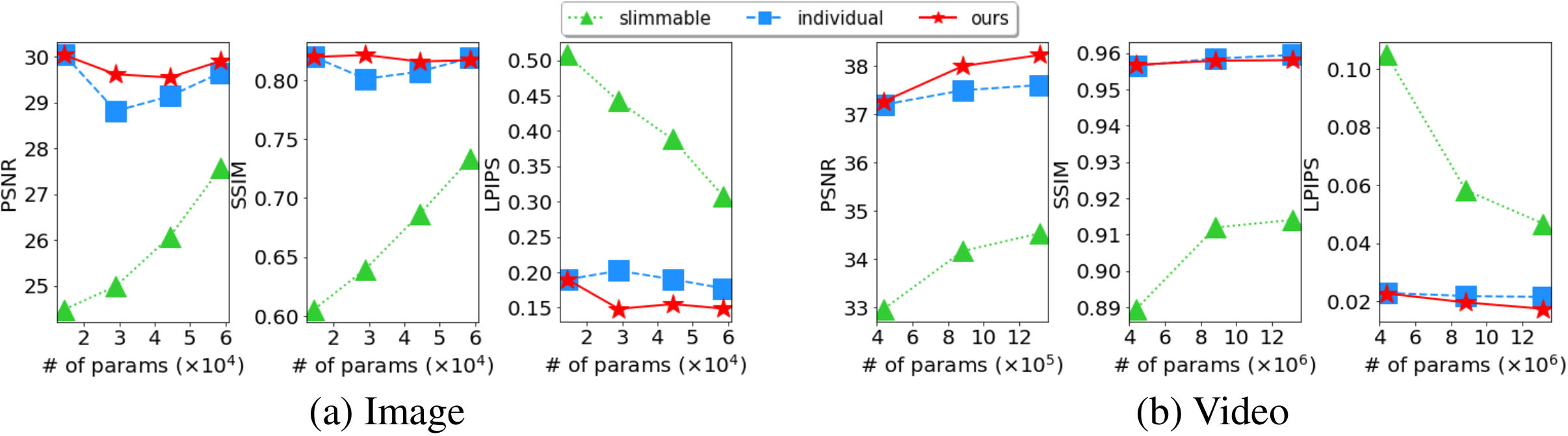}
  \caption{(a) Averaged performance of spatial growing of 8 Kodak images.
  (b) Averaged performance of temporal growing of 7 UVG videos.}
  \label{fig:spatial_temporal_result}
\end{figure*}

\subsection{Spatial and temporal growing}

\subsubsection{Images}
\label{sec:image_spatial}
We trained on 8 images in the Kodak dataset for spatial growing.
The network size grows three times and a total of four sub-networks represent spatially growing (horizontal direction) images.
We compared to baselines described in Fig.~\ref{fig:baseline_description} (b).
For a fair comparison, model prediction is compared against the signal domain that contains the desired signal in the evaluation stage.
The quantitative result in Fig.~\ref{fig:spatial_temporal_result} (a) shows that \textit{streamable (progressive)} models have higher representation power compared to the \textit{individual} models.
The performance drop along with the width increase implies that high-frequency details are concentrated in the center of the images.
Though the \textit{streamable (slimmable)} model falls behind, due to the sandwich rule~\cite{yu2019universally}, the representation power keeps increasing as the width grows.

\begin{wrapfigure}[12]{b}{0.6\textwidth}
\centering
  \vspace*{-0.6cm}
  \includegraphics[width=0.6\textwidth]{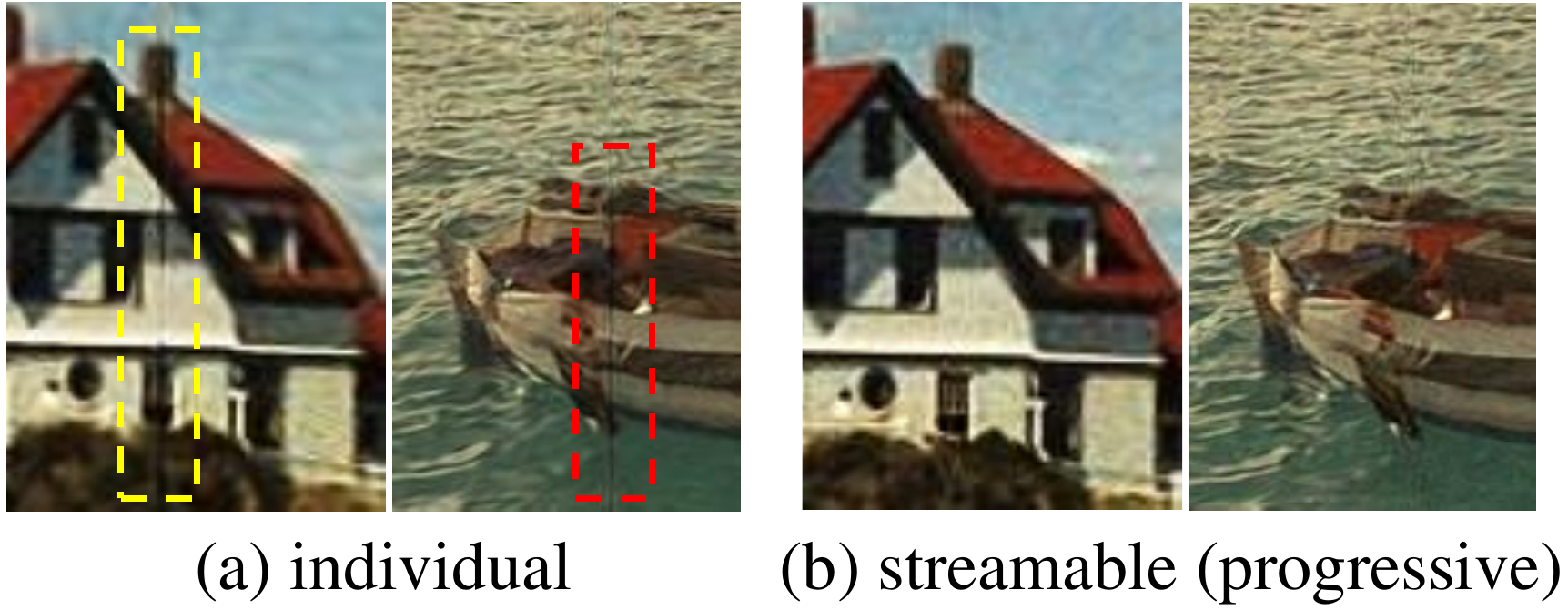}
  \caption{Individually trained model shows line artifacts while our model doesn't in spatial growing of images.}
  \label{fig:glitch}
\end{wrapfigure}

We also found that training independent models to spatially separate scenes yields line artifacts between each patch (Fig.~\ref{fig:glitch} (a)) because the entire scene is not taken into account during the training of each individual network.
Since these line artifacts are easily recognized, we believe our method has advantages over using individual models when learning spatially decomposed signals (Fig.~\ref{fig:glitch} (b)).
Seamlessly combining individually trained scenes is an active research area, and the recent Block-NeRF~\cite{tancik2022block} had to use additional alignment techniques, e.g., inverse distance weighting interpolation.

\subsubsection{Videos}
\label{sec:video_temporal}
We trained on 7 videos in UVG~\cite{mercat2020uvg} dataset, resized to 480$\times$270 for temporal growing.
The network size grows two times and a total of three sub-networks represent temporally growing videos.
Each sub-network reconstructs 8 frames, giving the largest network a total of 24 frames to reconstruct.
We compared to baselines described in Fig.~\ref{fig:baseline_description} (b).
As shown in Fig.~\ref{fig:spatial_temporal_result}, for the same network capacity, \textit{streamable (progressive)} models outperform the \textit{individual} models.

\subsection{Our Model Stabilizes the Training}
\label{stable}

\begin{figure*}[t]
  \includegraphics[width=\textwidth]{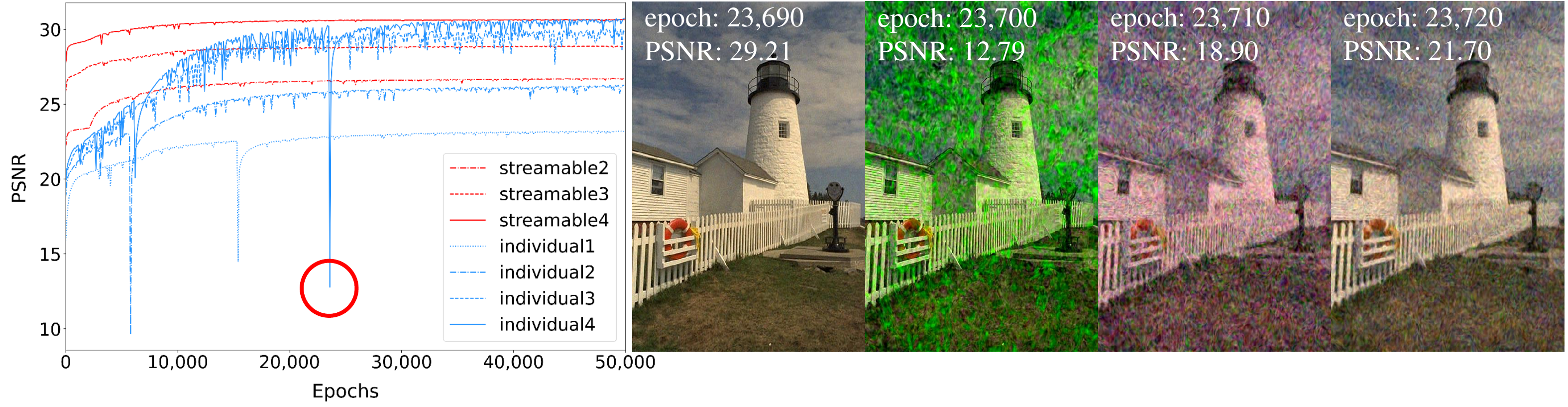}
  \vspace*{-7mm}
  \caption{PSNR curve and intermediate reconstructions of \textit{individual4} model during training.
  The \textit{individual} model's PSNR curve (blue) fluctuates during the entire training process while \textit{streamable (progressive)} (red) doesn't.}
  \vspace*{-2mm}
  \label{fig:loss_curve}
\end{figure*}

As shown in Fig.~\ref{fig:loss_curve}, the PSNR curves of \textit{individual} models drop significantly and then rebound during training, whereas those of \textit{streamable (progressive)} models show no abrupt changes. We reconstructed the predicted RGB values of the largest \textit{individual} model at the drop points and found severe artifacts on the images (Fig.~\ref{fig:loss_curve} right). This phenomenon is not limited to the example provided. When we trained the \textit{individual} models on 24 Kodak images, we found 19 similar cases. This suggests that the \textit{streamable (progressive)} model can maintain the training process stable.

Since high-frequency components are sensitive to the perturbations in network parameters~\cite{rahaman2019spectral}, fine-tuning the weights is required for a neural network to represent the high-frequency parts. Updating the entire parameters will significantly change the output signal and the loss value. This might be one reason why the individual model's PSNR curve fluctuates a lot. On the other hand, our progressive training method freezes pre-trained sub-network parameters and only updates the newly added weights. Updating partial weights results in small changes in the network output and encourages stable fine-tuning.




\section{Limitation and Discussion}
Visiting each width during the training stage is unavoidable if a neural network is to be runnable at varied architectures, and it requires a demanding training time.
There are several suggested parameter initialization methods on dynamically growing networks for fast convergence~\cite{wu2020firefly, wu2019splitting, evci2022gradmax}.
Further study on optimization and initialization techniques to improve the training dynamics can make our model more applicable for real-world tasks.

Conventional MLPs used in neural fields are black box models so that we cannot analyze how exactly the model reconstructs the desired signal.
The predicted signal of our model can be decoupled into partial reconstructions and the additive sense of predicting the output can give interpretability to the general machine learning and deep learning models~\cite{hastie2017generalized, agarwal2021neural}.
We believe our work can also bring insight into the interpretability of neural fields and encourage further research on neural field architectures in the community.



\section{Conclusion}
We showed the possibility of decoding neural signal representation in a streamline.
Our model is applicable to various signals such as images, videos, and 3D shapes which can vary in quality or grow spatially/temporally.
Compared with individually trained models, the streamable neural field has similar or higher representation power, along with efficient memory cost by utilizing parameter sharing.
Without re-training multiple individual models, our model can dynamically grow during the training procedure to search for optimal network capacity.
We have also shown that partially training the model weights stabilizes training.

\subsubsection{Acknowledgements}
This research was supported by the Ministry of Science and ICT (MSIT) of Korea, under the National Research Foundation (NRF) grant (2021R1F1A1061259), Institute of Information and Communication Technology Planning Evaluation (IITP) grants for the AI Graduate School program (IITP-2019-0-00421) and Artificial Intelligence Innovation Hub program (2021-0-02068).

\clearpage
%
%
\bibliographystyle{splncs04}
\bibliography{egbib}

\clearpage
\renewcommand\thesection{\Alph{section}}
\setcounter{section}{0}

\section{Ablation Studies}
\subsubsection{\textit{Weight initialization}}
\begin{figure*}[ht]
  \includegraphics[width=\textwidth]{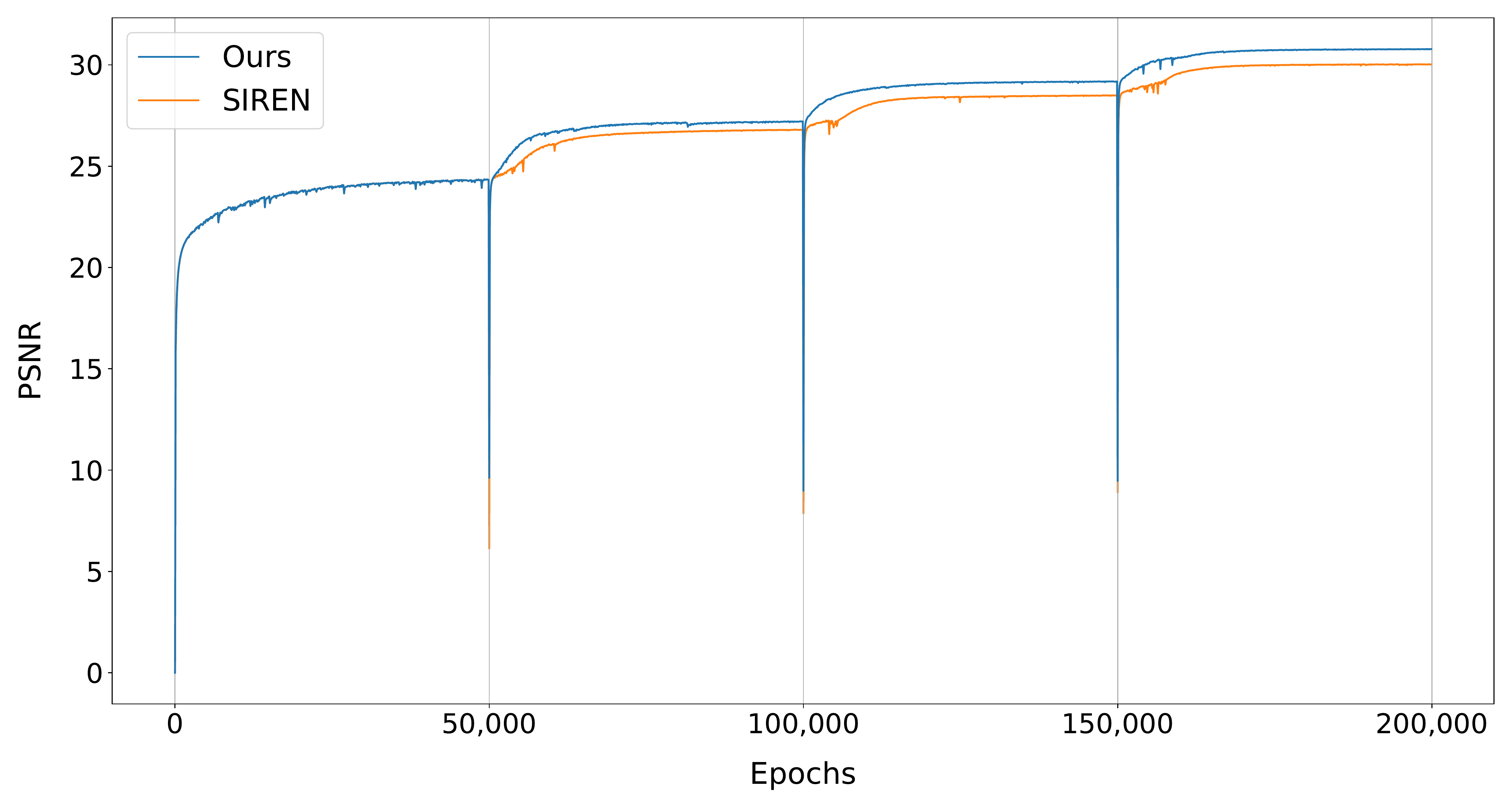}
  \caption{Averaged PSNR curve of two distinct weight initialization methods.
  We made the networks grow at every 50,000 epochs.
  Our proposed initialization makes the model converge at higher PSNR.}
  \label{fig:loss_curve_init}
\end{figure*}

\begin{figure*}[ht]
  \includegraphics[width=\textwidth]{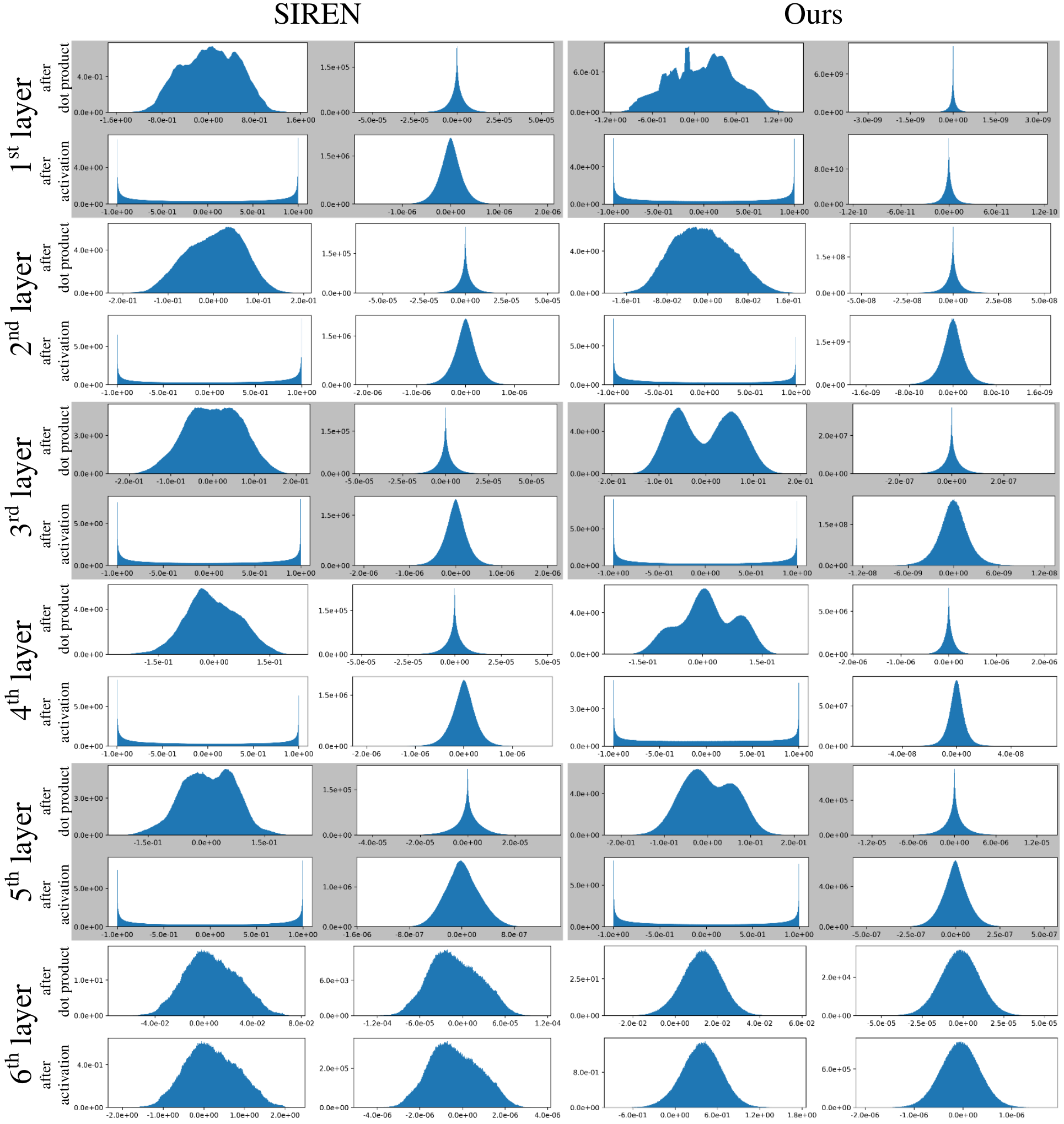}
  \caption{Activation and gradient statistics of SIREN and our initialization after 5 training epochs.
  Along with SIREN, our initialization method also preserves the distribution.
  Note that our model has no activation on the output layer.}
  \label{fig:distribution}
\end{figure*}

In this section, we demonstrate the effectiveness of our proposed initialization method.
While leaving the lateral connection as SIREN initialization~\cite{sitzmann2020implicit}, our method assigns zero to other newly added weights.
The newly added zero weights encourage the larger sub-network to start training when the network output is close to the output of the smaller pre-trained sub-network.
We trained two same \textit{streamable (progressive)} models on every image in the Kodak dataset but initialized the newly added weights differently.
Fig.~\ref{fig:loss_curve_init} shows that our initialization method supports faster convergence.
Also, we empirically found that our method converges at a higher PSNR compared to the SIREN initialization.
Furthermore, careful weight initialization is crucial when using sine activation to maintain the activation and gradient distribution at each layer~\cite{sitzmann2020implicit}.
Results shown in Fig.~\ref{fig:distribution} imply that adopting our initialization method does not harm the SIREN activation and gradient statistics.
\begin{figure*}[ht]
  \includegraphics[width=\textwidth]{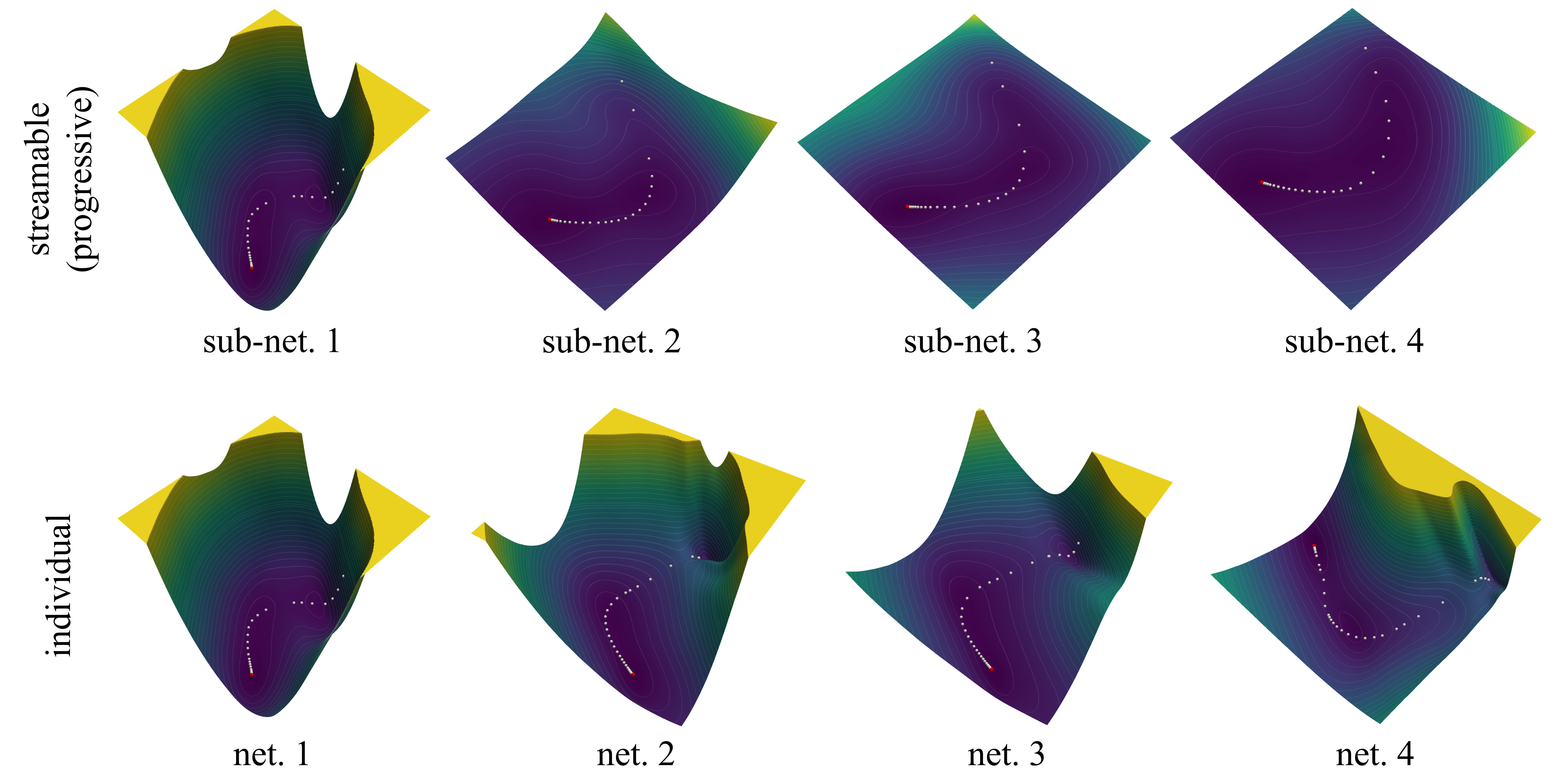}
  \caption{Visualization of loss surface and training trajectory.
  The \textit{individual} models' surface is more complex than that of \textit{streamable (progressive)} models.
  Each model is trained on Kodak image 19.
  }
  \label{fig:landscape}
\end{figure*}

\subsubsection{\textit{Training stability}}
We visualized the optimization trajectories of a \textit{streamable (progressive)} and an \textit{individual} model on 3D loss surfaces to better understand how our model keeps the training process stable.
Following Li et al.~\cite{li2018visualizing}, we applied PCA to matrix $M = [\theta_0 - \theta_n; \dots; \theta_{n-1} - \theta_n]$ and used the two leading principal components as direction vectors: $\delta$ and $\eta$.
Note that $\theta_i$ denotes network parameters at epoch $i$~\cite{li2018visualizing}.
One can plot a loss surface by moving the final parameters along the two directions and computing loss values.
More formally, we plot a function $f(\alpha, \beta)=L(\theta_n + \alpha\delta + \beta\eta)$, where $\alpha$ and $\beta$ are step sizes.
To plot the trajectories, we sampled the parameters every 5 epochs, computed the loss values, and projected them onto the loss surfaces.


As shown in the first column of Fig.~\ref{fig:landscape}, the loss surface of each model's smallest network has no extreme non-convexities.
However, the trajectories of the \textit{individual} model's larger networks traverse through non-convex regions (the second row), while the larger networks of the \textit{streamable (progressive)} model search for minima predominantly in low-loss regions (the first row).
This suggests that the knowledge (i.e., parameters) learned by the smaller sub-network gives a good initial point.
Since the larger network starts from a point where the loss value is already low, it converges fast and maintains the training process stable.


\section{Training Details and Additional Results}
We provide training details and additional results that are not specified in the main paper.
We show the metric values of each plot in the main paper along with qualitative results.
For every table, bold faces represent the best metric of each sub-network.
\subsubsection{\textit{Spectral growing}}
For 1D sinusoidal function reconstruction (section~\ref{sec:1d_recon}), we manufactured a target function $f(x) = \sum_{i=1}^{10}\sin{(2\pi k_ix + \phi_i)}$.
We set $k_i\in\{5, 10, ..., 50\}$, $\phi_i\sim U(0, 2\pi)$ and sampled $x$ uniformly over $[0, 1]$.
We used an MLP of three hidden layers with sine activations~\cite{sitzmann2020implicit}. We trained using Adam optimizer with a learning rate of $10^{-4}$.
Starting with a width of 10 (the channel size of hidden layers), we gradually increased the size up to 40.
Each sub-network was trained for 150 epochs.
For the image experiment (section~\ref{sec:spectral_image}), each model has five layers with sine activations and no activation on the final layer.
We used Adam~\cite{kingma2014adam} with a learning rate of $2\times10^{-4}$ and trained for 50,000 epochs.
For the SDF experiment (section~\ref{sec:spectral_sdf}), We used a learning rate of $10^{-4}$ and trained for 15,000 epochs.
The same loss function described in~\cite{sitzmann2020implicit} is used.
Other configurations are the same as the image experiment.

\begin{figure*}[ht]
  \includegraphics[width=\textwidth]{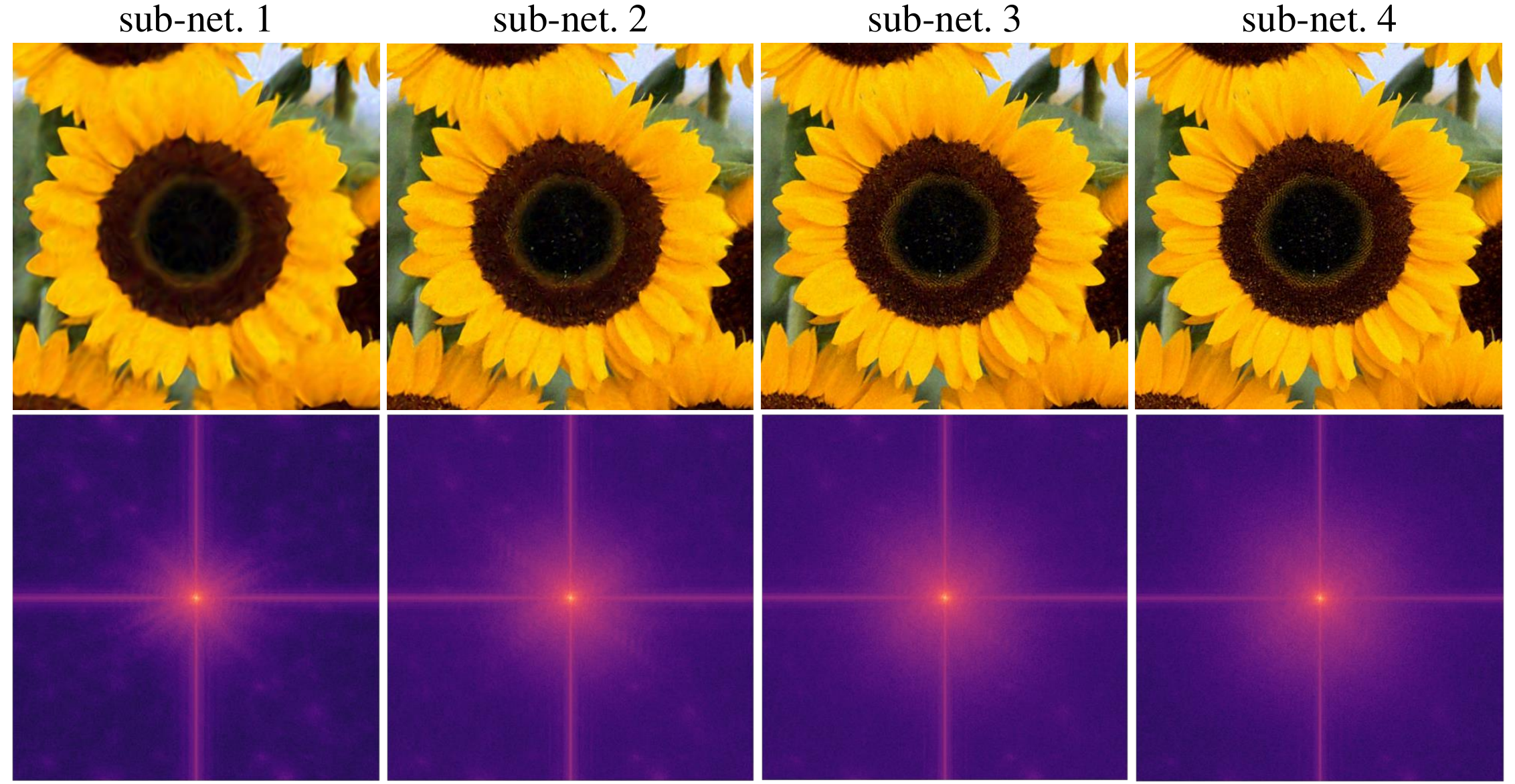}
  \caption{The frequency spectrum of spectral growing images represented by \textit{streamable (progressive)} neural fields.
  As the network grows, high-frequency components are gradually reconstructed (darker is smaller).
  }
  \label{fig:sunflower}
\end{figure*}

\begin{figure*}[ht]
  \includegraphics[width=\textwidth]{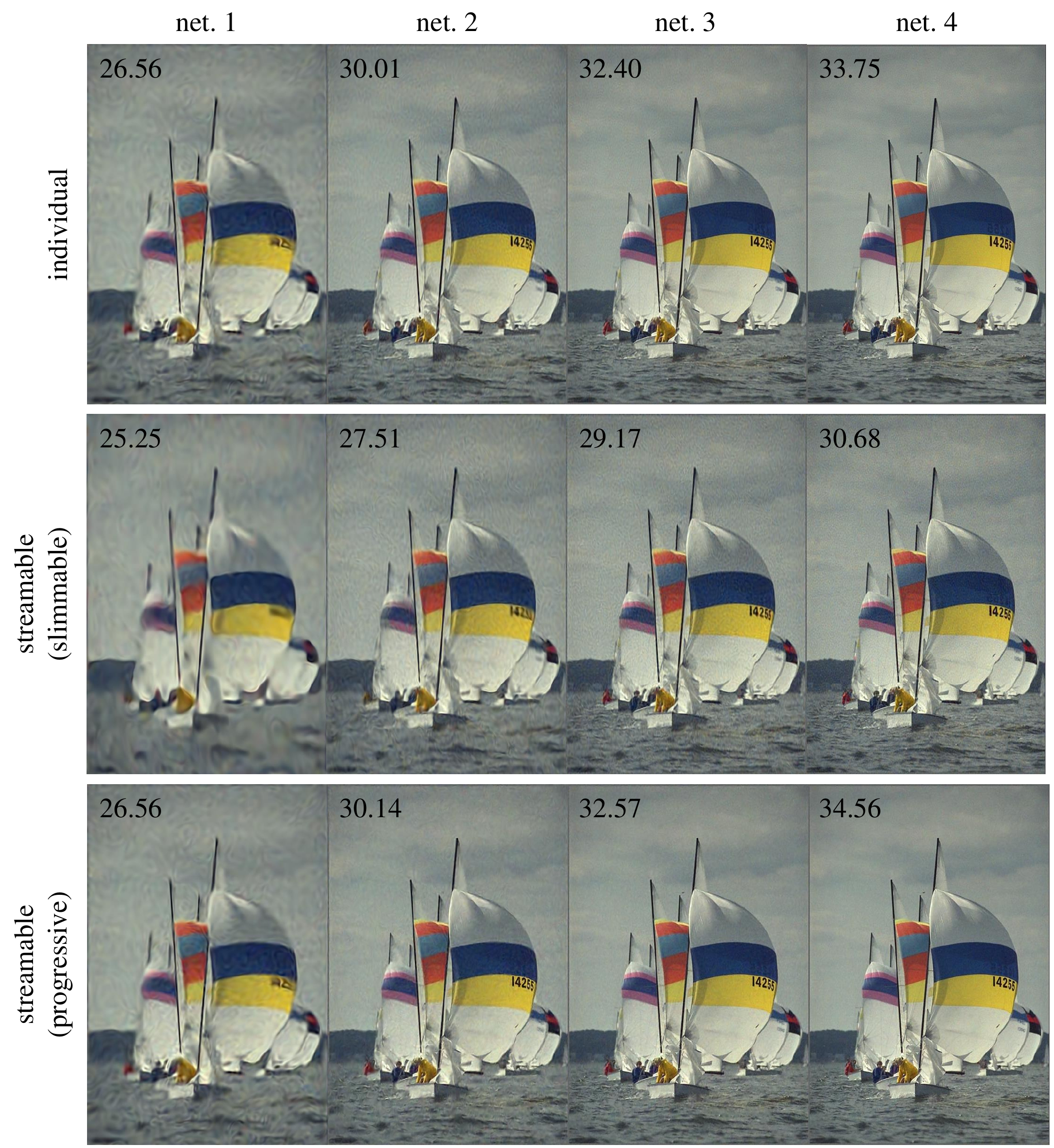}
  \caption{Qualitative results and PSNRs of spectral growing on Kodak image 9.
  \textit{Streamable (progressive)} neural field has the same or higher representation power compared to individually trained models.}
  \label{fig:spectral_image_supple}
\end{figure*}

\begin{figure*}[ht]
  \centering
  \includegraphics[width=4.64in]{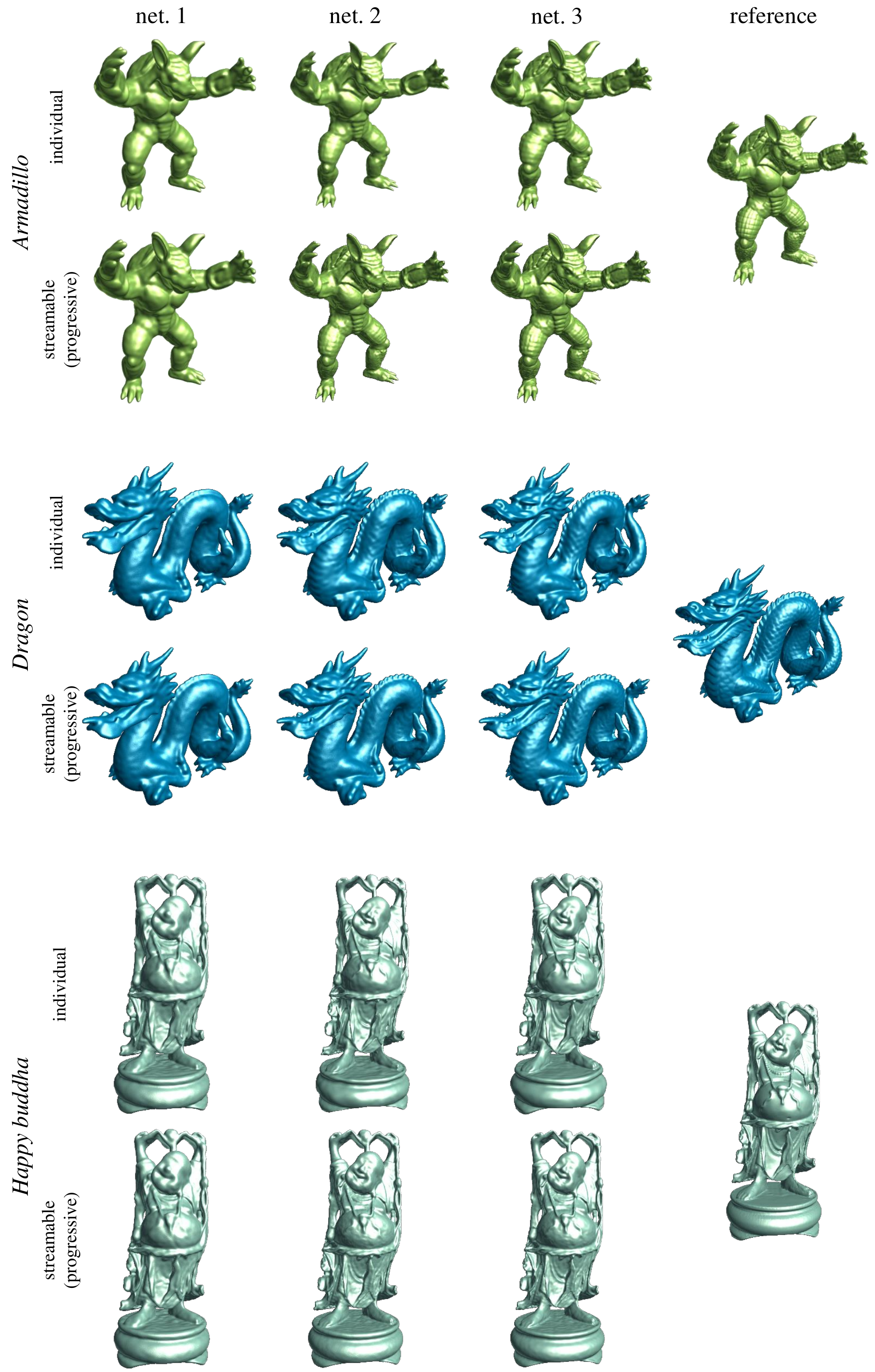}
  \caption{Qualitative results of spectral growing on 3D shapes.}
  \label{fig:sdf_supple}
\end{figure*}

Table.~\ref{tab:spectral_image} and Table.~\ref{tab:spectral_sdf} shows the quantitative results of spectral growing of images and 3D shapes respectively.
Qualitative results are depicted in Fig.~\ref{fig:spectral_image_supple} and Fig.~\ref{fig:sdf_supple}.
To show the growing frequency more directly, we fit an $800\times800$ sunflower image illustrated in Fig.~\ref{fig:sunflower}.
The second row shows the frequency domain of images reconstructed by \textit{streamable (progressive)} neural fields.
A larger network reconstructs higher frequency components that are not captured by smaller sub-networks.
We used the same network architecture and training settings used in Kodak experiments.
\subsubsection{\textit{Spatial/temporal growing}}
For the image spatial growing experiment (section~\ref{sec:image_spatial}), we used the same configurations used in the image spectral growing experiment.
For the video temporal growing experiment (section~\ref{sec:video_temporal}), we used a six-layer MLP and a learning rate of $10^{-4}$.
Other configurations are the same as the image experiment.

\begin{figure*}[ht]
  \centering
  \includegraphics[width=4.76in]{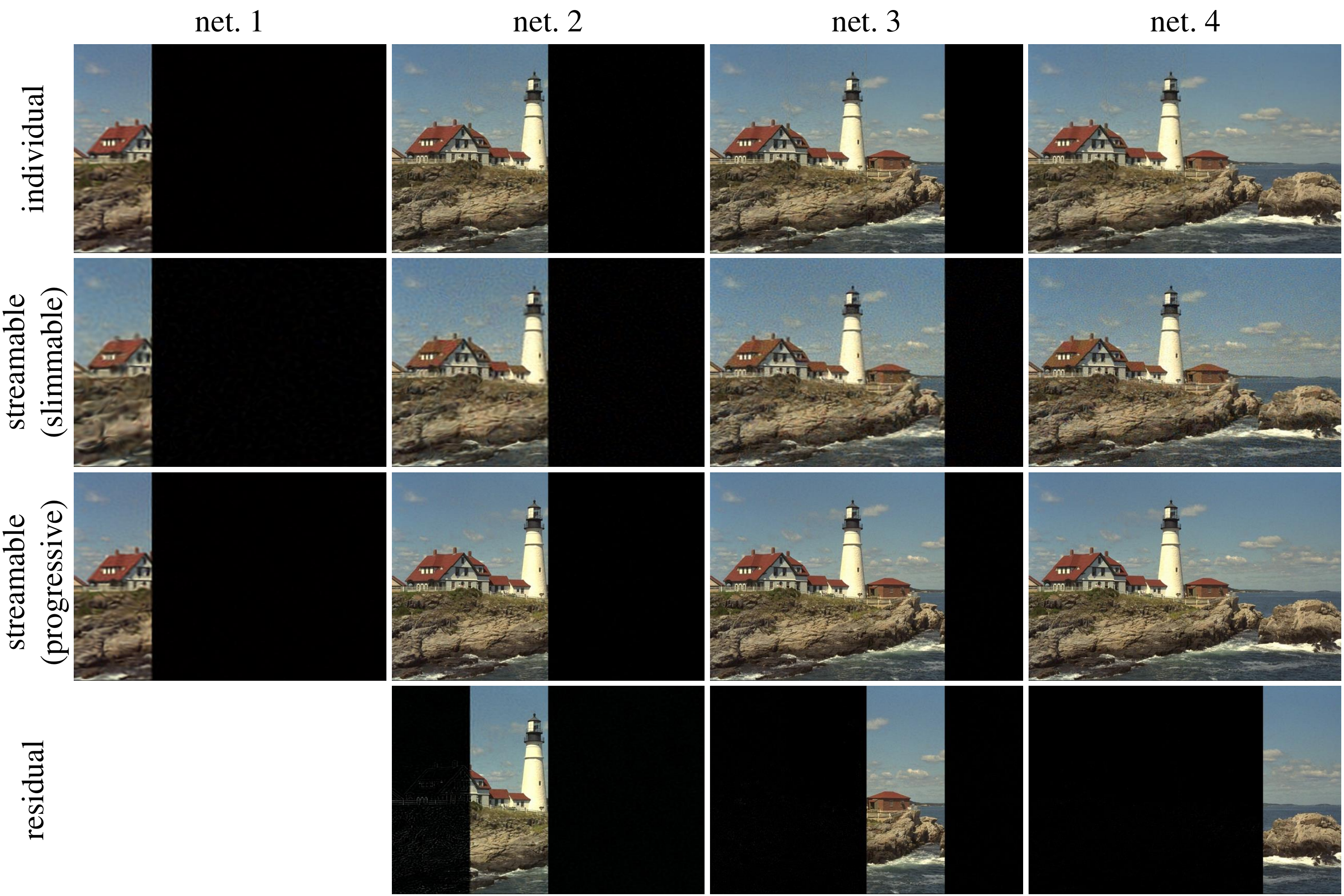}
  \vspace*{-4mm}
  \caption{Qualitative result of spatial growing on Kodak image 21.}
  \label{fig:spatial_supple}
\end{figure*}

\begin{figure*}[ht]
  \centering
  \includegraphics[width=\textwidth]{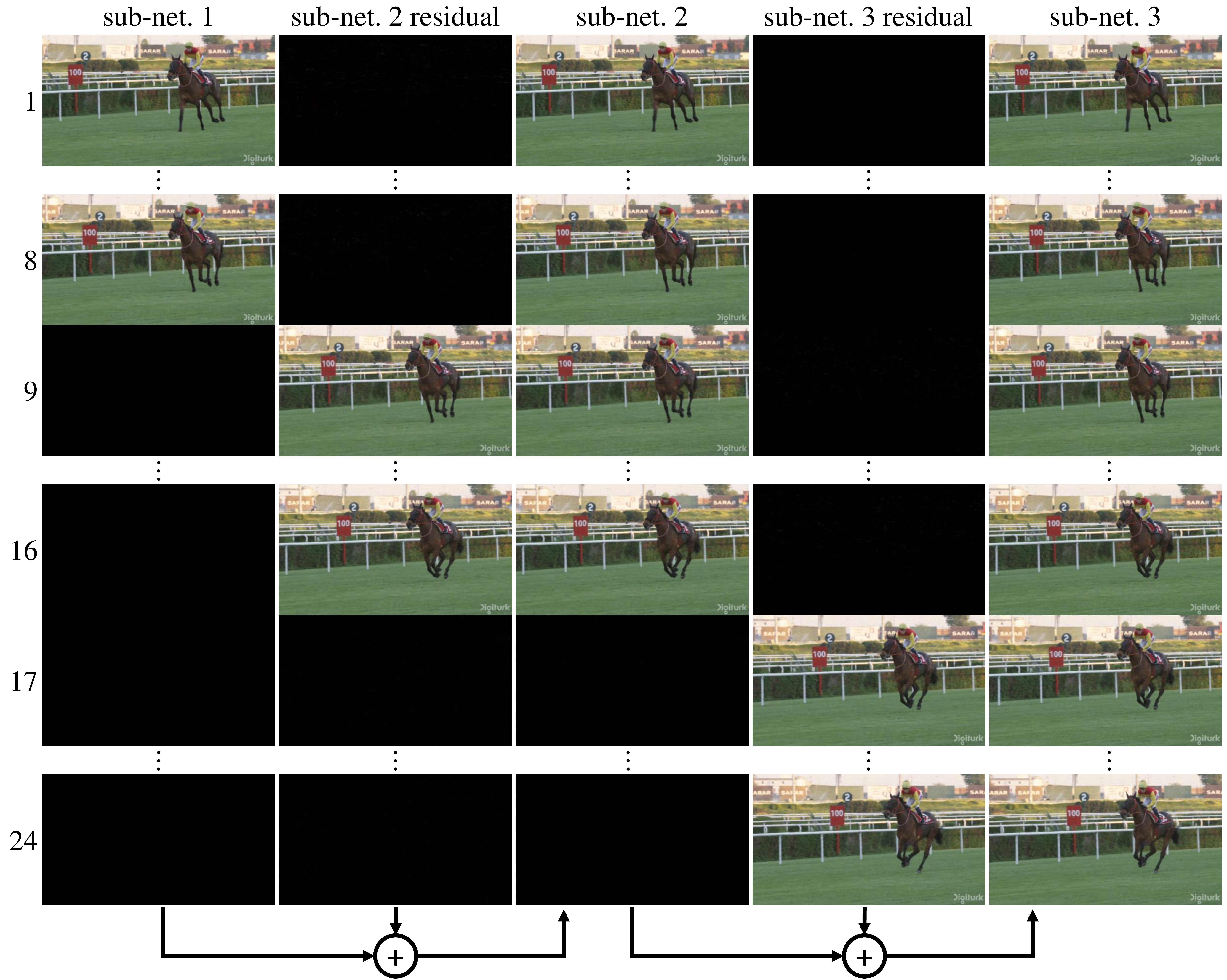}
  \vspace*{-8mm}
  \caption{Qualitative result of temporal growing on video (\textit{Jockey}).
  A larger sub-network reconstructs the exact residual frames that are not represented by the smaller ones.
  }
  \label{fig:video_supple}
\end{figure*}

Quantitative results of baseline comparison experiments of spatial/temporal growing are shown in Table.~\ref{tab:spatial_image} and Table.~\ref{tab:temporal} respectively.
Qualitative results are depicted in Fig.~\ref{fig:spatial_supple} and Fig.~\ref{fig:video_supple}.
To obtain a longer temporal growing video, we fit the ``bikes sequence" video, which consists of $272\times640$ pixels and 240 frames using the same network architecture and training settings used in the UVG dataset~\cite{mercat2020uvg} experiments.
We provided the reconstructed video in \textit{.mp4} format to show the result.
\begin{table}[ht]
\centering
    \begin{tabular}{@{\hskip 0.1in}c@{\hskip 0.1in}|@{\hskip 0.1in}c@{\hskip 0.1in}|@{\hskip 0.1in}c@{\hskip 0.2in}c@{\hskip 0.2in}c@{\hskip 0.2in}c@{\hskip 0.1in}c}
    \multirow{2}{*}{sub-network} & \multirow{2}{*}{metrics} & \multirow{2}{*}{individual} & streamable & streamable \\
    & & & (slimmable~\cite{yu2019slimmable}) & (progressive) \\
    \hline
    \multirow{4}{1em}{1} & params. & 3,903 & 3,903 & 3,903 \\ 
    & PSNR$\uparrow$ & 24.39 & 23.72 & 24.39 \\ 
    & SSIM$\uparrow$ & 0.597 & 0.578 & 0.597 \\
    & LPIPS$\downarrow$ & 0.559 & 0.582 & 0.559 \\ 
    \hline
    \multirow{4}{1em}{2} & params. & 15,003 & 15,003 & 15,057 \\ 
    & PSNR$\uparrow$ & 27.22 & 25.64 & \textbf{27.30} \\ 
    & SSIM$\uparrow$ & 0.701 & 0.643 & \textbf{0.704} \\
    & LPIPS$\downarrow$ & 0.380 & 0.462 & \textbf{0.298} \\ 
    \hline
    \multirow{4}{1em}{3} & params. & 33,303 & 33,303 & 33,507 \\ 
    & PSNR$\uparrow$ & \textbf{29.78} & 27.29 & 29.29 \\ 
    & SSIM$\uparrow$ & \textbf{0.795} & 0.704 & 0.775 \\
    & LPIPS$\downarrow$ & 0.246 & 0.372 & \textbf{0.190} \\ 
    \hline
    \multirow{4}{1em}{4} & params. & 58,803 & 58,803 & 58,681 \\ 
    & PSNR$\uparrow$ & \textbf{31.19} & 29.19 & 30.91 \\ 
    & SSIM$\uparrow$ & \textbf{0.836} & 0.770 & 0.825 \\
    & LPIPS$\downarrow$ & 0.149 & 0.271 & \textbf{0.129} \\ 
    \hline
    \end{tabular}
    \vspace*{2mm}
    \caption{Quantitative results of spectral growing on 24 Kodak images.}
    \label{tab:spectral_image}
\end{table}
\begin{table}[b]
\centering
    \begin{tabular}{@{\hskip 0.1in}c@{\hskip 0.1in}|@{\hskip 0.1in}c@{\hskip 0.1in}|@{\hskip 0.1in}c@{\hskip 0.2in}c@{\hskip 0.2in}c@{\hskip 0.2in}c}
    \multirow{2}{*}{sub-network} & \multirow{2}{*}{shapes} & \multirow{2}{*}{individual} & streamable \\
    & & & (progressive) \\
    \hline
    \multirow{4}{1em}{1} & params. & 66,897 & 66,897 \\ 
    & \textit{armadillo} & 0.321 & 0.321 \\ 
    & \textit{dragon} & 0.104 & 0.104 \\
    & \textit{happy buddha} & 0.109 & 0.109 \\ 
    \hline
    \multirow{4}{1em}{2} & params. & 133,873 & 133,981 \\ 
    & \textit{armadillo} & 0.137 & \textbf{0.032} \\ 
    & \textit{dragon} & 0.128 & \textbf{0.030} \\
    & \textit{happy buddha} & 0.054 & \textbf{0.019} \\ 
    \hline
    \multirow{4}{1em}{3} & params. & 198,388 & 198,657 \\ 
    & \textit{armadillo} & 0.059 & \textbf{0.023} \\ 
    & \textit{dragon} & 0.056 & \textbf{0.025} \\
    & \textit{happy buddha} & 0.042 & \textbf{0.018} \\ 
    \hline
    \end{tabular}
    \vspace*{2mm}
    \caption{Chamfer distance of spectral growing 3D shapes.}
    \label{tab:spectral_sdf}
\end{table}
\begin{table}[ht]
\centering
    \begin{tabular}{@{\hskip 0.1in}c@{\hskip 0.1in}|@{\hskip 0.1in}c@{\hskip 0.1in}|@{\hskip 0.1in}c@{\hskip 0.2in}c@{\hskip 0.2in}c@{\hskip 0.2in}c@{\hskip 0.1in}c}
    \multirow{2}{*}{sub-network} & \multirow{2}{*}{metrics} & \multirow{2}{*}{individual} & streamable & streamable \\
    & & & (slimmable~\cite{yu2019slimmable}) & (progressive) \\
    \hline
    \multirow{4}{1em}{1} & params. & 14,517 & 14,517 & 14,517 \\ 
    & PSNR$\uparrow$ & 30.04 & 24.49 & 30.04 \\ 
    & SSIM$\uparrow$ & 0.820 & 0.606 & 0.820 \\
    & LPIPS$\downarrow$ & 0.190 & 0.508 & 0.190 \\ 
    \hline
    \multirow{4}{1em}{2} & params. & 29,034 & 29,067 & 29,095 \\ 
    & PSNR$\uparrow$ & 28.82 & 24.99 & \textbf{29.62} \\ 
    & SSIM$\uparrow$ & 0.801 & 0.640 & \textbf{0.821} \\
    & LPIPS$\downarrow$ & 0.202 & 0.442 & \textbf{0.148} \\ 
    \hline
    \multirow{4}{1em}{3} & params. & 43,551 & 44,307 & 44,515 \\ 
    & PSNR$\uparrow$ & 29.14 & 26.07 & \textbf{29.55} \\ 
    & SSIM$\uparrow$ & 0.807 & 0.686 & \textbf{0.816} \\
    & LPIPS$\downarrow$ & 0.190 & 0.388 & \textbf{0.155} \\ 
    \hline
    \multirow{4}{1em}{4} & params. & 58,068 & 58,803 & 58,453 \\ 
    & PSNR$\uparrow$ & 29.65 & 27.58 & \textbf{29.91} \\ 
    & SSIM$\uparrow$ & \textbf{0.819} & 0.734 & 0.817 \\
    & LPIPS$\downarrow$ & 0.177 & 0.307 & \textbf{0.149} \\ 
    \hline
    \end{tabular}
    \vspace*{2mm}
    \caption{Quantitative results of spatial growing on 8 Kodak images.}
    \label{tab:spatial_image}
\end{table}
\begin{table}[ht]
\centering
    \begin{tabular}{@{\hskip 0.1in}c@{\hskip 0.1in}|@{\hskip 0.1in}c@{\hskip 0.1in}|@{\hskip 0.1in}c@{\hskip 0.2in}c@{\hskip 0.2in}c@{\hskip 0.2in}c@{\hskip 0.1in}c}
    \multirow{2}{*}{sub-network} & \multirow{2}{*}{metrics} & \multirow{2}{*}{individual} & streamable & streamable \\
    & & & (slimmable~\cite{yu2019slimmable}) & (progressive) \\
    \hline
    \multirow{4}{1em}{1} & params. & 441,635 & 441,635 & 441,635 \\ 
    & PSNR$\uparrow$ & 37.20 & 32.97 & 37.27 \\ 
    & SSIM$\uparrow$ & 0.957 & 0.890 & 0.957 \\
    & LPIPS$\downarrow$ & 0.023 & 0.105 & 0.023 \\ 
    \hline
    \multirow{4}{1em}{2} & params. & 883,270 & 882,836 & 882,116 \\ 
    & PSNR$\uparrow$ & 37.49 & 34.17 & \textbf{37.99} \\ 
    & SSIM$\uparrow$ & \textbf{0.959} & 0.912 & 0.958 \\
    & LPIPS$\downarrow$ & 0.022 & 0.058 & \textbf{0.020} \\ 
    \hline
    \multirow{4}{1em}{3} & params. & 1,324,905 & 1,316,867 & 1,317,096 \\ 
    & PSNR$\uparrow$ & 37.60 & 34.54 & \textbf{38.22} \\ 
    & SSIM$\uparrow$ & \textbf{0.960} & 0.914 & 0.958 \\
    & LPIPS$\downarrow$ & 0.022 & 0.047 & \textbf{0.017} \\ 
    \hline
    \end{tabular}
    \vspace*{2mm}
    \caption{Quantitative results of temporal growing on 7 UVG videos.}
    \label{tab:temporal}
\end{table}
\vspace*{-8mm}
\clearpage

\end{document}